\documentclass[10pt,journal,compsoc]{IEEEtran}

\usepackage{subfigure}
\usepackage{amsmath,amsfonts}

\usepackage{mathrsfs}

\usepackage[linesnumbered,ruled,vlined]{algorithm2e}
\usepackage{array}
\usepackage[caption=false,font=normalsize,labelfont=sf,textfont=sf]{subfig}
\usepackage{textcomp}
\usepackage{stfloats}
\usepackage{url}
\usepackage{verbatim}
\usepackage{graphicx}
\usepackage{cite}
\usepackage{subfig}
\usepackage{indentfirst}
\usepackage{graphicx}   

\usepackage{refcount}

\usepackage{amsmath}   
\usepackage{cuted}     
\usepackage{lipsum}    

\usepackage{booktabs}
\usepackage{tabularx, multirow, amssymb, textcomp, array}
\usepackage{colortbl} 

\usepackage{titlesec}

\begin{document}

\title{Task Assignment and Exploration Optimization for Low Altitude UAV Rescue via Generative AI Enhanced Multi-agent Reinforcement Learning}

\author{Xin Tang, Qian Chen, Wenjie Weng, Chao Jin, Zhang Liu, Jiacheng Wang,
Geng Sun,~\IEEEmembership{Senior Member,~IEEE,}
        Xiaohuan Li,~\IEEEmembership{Member,~IEEE,}
        Dusit Niyato,~\IEEEmembership{Fellow,~IEEE}

\IEEEcompsocitemizethanks{

\IEEEcompsocthanksitem Xin Tang is with the Guangxi University Key Laboratory of Intelligent Networking and Scenario System (School of Information and Communication, Guilin University of Electronic Technology(GUET)), Guilin, 541004, China, and also with the College of Computing and Data Science, Nanyang Technological University, Singapore 639798 (e-mail: tangx@mails.guet.edu.cn).
\IEEEcompsocthanksitem Qian Chen is with the School of Architecture and Transportation Engineering, GUET, Guilin, 541004, China (e-mail: chenqian@mails.guet.edu.cn).
\IEEEcompsocthanksitem Wenjie Weng, Chao Jin and Xiaohuan Li are with the Guangxi University Key Laboratory of Intelligent Networking and Scenario System (School of Information and Communication, GUET), Guilin, 541004, China, and also with National Engineering Laboratory for Comprehensive Transportation Big Data Application Technology (Guangxi), Nanning, 530001, China (e-mails: wwjdzsyx@163.com; kingchao2025@163.com; lxhguet@guet.edu.cn).
\IEEEcompsocthanksitem Zhang Liu is with the Department of Informatics and Communication Engineering, Xiamen University, Fujian, 361102, China (e-mail: zhangliu@stu.xmu.edu.cn).
\IEEEcompsocthanksitem Geng Sun is with the College of Computer Science and Technology, Jilin University, Changchun 130012, China, and with Key Laboratory of Symbolic Computation and Knowledge Engineering of Ministry of Education, Jilin University, Changchun 130012, China (e-mail: sungeng@jlu.edu.cn).
\IEEEcompsocthanksitem Jiacheng Wang and Dusit Niyato are with the College of Computing and Data Science, Nanyang Technological University, Singapore 639798 (e-mails: jiacheng.wang@ntu.edu.sg; dniyato@ntu.edu.sg).}
\thanks{This work was supported in part by the Guangxi Natural Science Foundation of China under Grant 2025GXNSFAA069687, in part by the National Natural Science Foundation of China under Grant U22A2054, and in part by the Graduate Study Abroad Program of GUET under Grant GDYX2024001. {\itshape (Corresponding author: Xiaohuan Li).}}

}

\markboth{IEEE Transactions on Mobile Computing,~Vol.~X, No.~X, X~2025}%
{Shell \MakeLowercase{\textit{et al.}}: Bare Advanced Demo of IEEEtran.cls for IEEE Computer Society Journals}

\IEEEtitleabstractindextext{%
\begin{abstract}
The integration of emerging uncrewed aerial vehicle (UAV) with artificial intelligence (AI) and ground-embedded robots (GERs) has transformed emergency rescue operations in unknown environments. However, the high computational demands of such missions often exceed the capacity of a single UAV, making it difficult for the system to continuously and stably provide high-level services. To address these challenges, this paper proposes a novel cooperation framework involving UAVs, GERs, and airships. This framework enables resource pooling through UAV-to-GER (U2G) and UAV-to-airship (U2A) communications, providing computing services for UAV offloaded tasks. Specifically, we formulate the multi-objective optimization problem of task assignment and exploration optimization in UAVs as a dynamic long-term optimization problem. Our objective is to minimize task completion time and energy consumption while ensuring system stability over time. To achieve this, we first employ the Lyapunov optimization method to transform the original problem, with stability constraints, into a per-slot deterministic problem. We then propose an algorithm named HG-MADDPG, which combines the Hungarian algorithm with a generative diffusion model (GDM)-based multi-agent deep deterministic policy gradient (MADDPG) approach, to jointly optimize exploration and task assignment decisions. In HG-MADDPG, we first introduce the Hungarian algorithm as a method for exploration area selection, enhancing UAV efficiency in interacting with the environment. We then innovatively integrate the GDM and multi-agent deep deterministic policy gradient (MADDPG) to optimize task assignment decisions, such as task offloading and resource allocation. Simulation results demonstrate the effectiveness of the proposed approach, with significant improvements in task offloading efficiency, latency reduction, and system stability compared to baseline methods.
\end{abstract}

}

\maketitle

\IEEEdisplaynontitleabstractindextext

\IEEEpeerreviewmaketitle

\ifCLASSOPTIONcompsoc
\IEEEraisesectionheading{\section{Introduction}\label{sec:introduction}}
\else
\section{Introduction}
\label{sec:introduction}
\fi

\IEEEPARstart{I}{nformation} collection and target detection from the disaster area is crucial to assess the situation and develop a suitable rescue plan \cite{fang2025prioritized}. When ground transportation is interrupted by a disaster, it becomes difficult for human rescuers to enter the affected area. The existing method for rescuing in disaster areas involves using either ground vehicles or aerial robots that operate independently to gather data and transmit rescue information back to a ground commander \cite{karaman2024enhancing, he2025advancing}. Furthermore, ground-embedded robots (GERs) are used to explore for life forms and for environmental monitoring, while low-altitude uncrewed aerial vehicles (UAVs) play a significant role in post-disaster search and rescue activities. Emerging UAVs, combined with artificial intelligence (AI), have revolutionized the way emergency rescue is handled \cite{liu2024generative, tang2024digital}. The UAVs offer several advantages in emergency rescue and disaster response. First, they can quickly access areas that are otherwise unreachable due to collapsed infrastructure or hazardous conditions, enabling rapid situational assessment. Second, UAVs equipped with high-resolution cameras and sensors can provide real-time aerial imagery and environmental data, greatly enhancing decision-making efficiency for rescue teams. Third, their flexibility in deployment allows for coordinated missions across wide areas, improving coverage and responsiveness. These capabilities make UAVs a valuable asset in time-sensitive and high-risk rescue operations. Despite the advantages of low-altitude UAVs for emergency rescue, there are still substantial challenges that need to be addressed.


{\itshape Challenge 1: While existing frameworks utilize UAVs for low-altitude operations, they often neglect the heterogeneous capabilities and practical constraints of collaborating computing nodes in dynamic rescue scenarios.}
The absence of a unified framework that jointly considers task offloading prioritization, obstacle-aware communication reliability, and resource availability limits the adaptability of rescue systems \cite{xiaohuan2020aggregate}. This motivates the development of a novel cooperation framework that dynamically coordinates multiple nodes while addressing real-world constraints such as intermittent connectivity and heterogeneous computational capacities \cite{li2024cloud}.

{\itshape Challenge 2: Existing approaches to computation offloading in UAV rescue scenarios fail to holistically optimize latency, energy consumption, and exploration efficiency under time-varying resource availability and environmental obstacles.}
The interdependencies among task assignment, energy constraints, and obstacle-induced communication disruptions create a complex trade-off space \cite{sun2024multi}. Moreover, ensuring long-term system stability while minimizing instantaneous task latency remains an open challenge \cite{liu2024dnn}. This necessitates an online optimization method capable of decomposing long-term objectives into real-time decisions while maintaining robustness against dynamic uncertainties.

{\itshape Challenge 3: Current multi-agent reinforcement learning (MARL)-based solutions for multi-agent coordination often overlook the impact of agent observation limitations and high-dimensional state-action spaces on strategy generation.}
In rescue scenarios, agents operate with partial observations due to obstacles and limited sensing ranges, which lead to suboptimal task assignment and exploration decisions \cite{sun2024generative}. Furthermore, the curse of dimensionality in multi-agent systems hinders efficient policy learning \cite{liu2024two}. These challenges highlight the need for a hybrid approach that combines low-complexity algorithms with AI techniques to reduce observation space complexity and enhance collaborative decision-making under partial observability.

Motivated by these considerations, we propose a novel computing task assignment method for UAVs that leverages a variety of GERs to provide computing offloading service. Given the dynamics of competition and cooperation between UAVs and GERs, we present a stable task assignment algorithm designed to optimally pair each UAV with the GER that best meets their demand and supply. Moreover, we formulate the task assignment and exploration optimization for UAVs as a mixed-integer nonlinear optimization problem. To address this problem, we employ an online algorithm that transforms the long-term optimization problem into a real-time, instantaneous optimization problem using the Lyapunov optimization method. Furthermore, a Hungarian algorithm and generative diffusion model (GDM)-based MARL method are proposed to solve the instantaneous optimization problem. The main contributions of this paper are summarized as follows:

\begin{itemize}
\item{
\textbf{Framework:} We propose a novel cooperation framework involving UAVs, GERs, and airships in low-altitude rescue scenarios, where computationally intensive tasks from UAVs can be offloaded to GERs or airships for more efficient execution. Specifically, UAVs prioritize direct assignment of tasks to GERs to minimize latency. An airship is engaged to handle offloaded tasks only when GERs lack sufficient computation resources. Moreover, in scenarios where obstacles obstruct communication between the ground control center and GERs, the UAVs can detect obstacles, facilitating reliable rescue.
}

\item{
\textbf{Multi-objective optimization (MOO):} We jointly optimize the task assignment, energy consumption and the exploration area selection problem for GERs and UAVs to minimize the completion latency of tasks. We employ an online algorithm that addresses the long-term optimization problem by converting the problem into a real-time, instantaneous optimization problem by using the Lyapunov optimization method. This strategy effectively decouples the minimization of long-term task completion time with stability constraints into a deterministic problem for each time slot.
}

\item{
\textbf{Solution:} To address the complex optimization problem described above, this paper models the MOO problem as a markov decision process (MDP) and proposes a method based on the Hungarian algorithm and GDM-based multiagent deep deterministic policy gradient (MADDPG), named HG-MADDPG. Specifically, we introduce a novel application of the Hungarian algorithm for exploration area selection, which reduces the dimensionality of the observation space. The  exploration process involves trajectory generation and obstacle detection. Additionally, we integrate the GDM and MADDPG to enhance the generative decision-making capability of the actor network, enabling the optimization of task assignment.
}

\item{
\textbf{Validation:} Extensive experiments are conducted to illustrate the substantial advantages of the proposed approach in comparison to baseline algorithms, including more stable task completion latency, lower energy consumption, and enhanced system stability.
}
\end{itemize}

The rest of the paper is organized as follows. The related works are reviewed in section \ref{sec:2}. In section \ref{sec:3}, we present the system model and optimization problem. Section \ref{sec:4} reformulates the optimization problem. Section \ref{sec:5} introduces the novel method based on the HG-MADDPG algorithm for task assignment and exploration optimization. Section \ref{sec:6} is dedicated to the simulation experiments and their analysis, while section \ref{sec:7} provides the concluding remarks.

\begin{table*}[t]
\centering
\caption{A $\checkmark$ in the table indicates that the performance metrics meet the specified criteria. \label{tab:table01}}
\begin{tabularx}{\textwidth}{|c|>{\centering\arraybackslash}X|c|c|c|c|c|c|c|}
\hline
\multirow{2}{*}{\begin{tabular}{@{}c@{}}Ref.\end{tabular}} &  
\multirow{2}{*}{\begin{tabular}{@{}c@{}}Task assignment \\ scenario\end{tabular}} & 
\multirow{2}{*}{\begin{tabular}{@{}c@{}}Task latency\end{tabular}} & 
\multirow{2}{*}{\begin{tabular}{@{}c@{}}Energy \\ consumption \end{tabular}} & 
\multirow{2}{*}{\begin{tabular}{@{}c@{}}UAV \\ trajectory\end{tabular}} & 
\multirow{2}{*}{\begin{tabular}{@{}c@{}}Task completion \\ rate\end{tabular}} &  
\multirow{2}{*}{\begin{tabular}{@{}c@{}}low-altitude \\ obstacle\end{tabular}} &  
\multirow{2}{*}{\begin{tabular}{@{}c@{}}Resource \\ visualization\end{tabular}} &
\multirow{2}{*}{\begin{tabular}{@{}c@{}}Stability\end{tabular}} \\
& & & & & & & & \\  \hline
\cite{sun2025task}  & \multirow{8}{*}{\begin{tabular}{@{}c@{}}GER to UAV\end{tabular}} & \checkmark & \checkmark & \checkmark & -- & -- &-- & -- \\ \cline{1-1} \cline{3-9}
\cite{azfar2025enhancing}  &  & \checkmark & \checkmark & \checkmark & -- & -- &\checkmark  & -- \\ \cline{1-1} \cline{3-9} 
\cite{xu2025ideas}  &  & \checkmark & \checkmark & \checkmark & -- & -- &--  & \checkmark \\ \cline{1-1} \cline{3-9}
\cite{zhao2025joint}  & & \checkmark & \checkmark & \checkmark & -- & -- &-- & -- \\ \cline{1-1} \cline{3-9}
\cite{xiao2025star}  &  & -- & \checkmark & \checkmark & -- & -- &--  & -- \\ \cline{1-1} \cline{3-9} 
\cite{zhou2025user}  & & \checkmark & \checkmark & \checkmark & -- & -- &-- & -- \\ \cline{1-1} \cline{3-9}
\cite{tang2023digital}  & & \checkmark & \checkmark & \checkmark & \checkmark & -- &-- & -- \\ \cline{1-1} \cline{3-9}
\cite{liu2023energy}  & &-- &\checkmark & \checkmark & -- & -- &-- &-- \\  \hline
\cite{hazarika2025generative} & \multirow{4}{*}{\begin{tabular}{@{}c@{}}UAV to UAV\end{tabular}} & \checkmark & \checkmark & -- & \checkmark & \checkmark & --  & -- \\ \cline{1-1} \cline{3-9}
\cite{tang2025dnn}  & & \checkmark & \checkmark & \checkmark & \checkmark & -- & --  & \checkmark \\ \cline{1-1} \cline{3-9}
\cite{sun2024all}  & & \checkmark &-- & -- & -- & -- &-- &\checkmark \\ \cline{1-1} \cline{3-9}
\cite{raivi2024jdaco}  & &\checkmark &-- & \checkmark & \checkmark & -- &-- &-- \\  \hline
\cite{sun2024joint} & \multirow{3}{*}{\begin{tabular}{@{}c@{}}UAV to GER\end{tabular}} &\checkmark & \checkmark & -- & -- & -- &-- &-- \\ \cline{1-1} \cline{3-9}
\cite{9696188}  & & \checkmark & \checkmark & \checkmark & -- & \checkmark &-- & -- \\ \cline{1-1} \cline{3-9}
\begin{tabular}{@{}c@{}}This  work \end{tabular}& & \checkmark & \checkmark & \checkmark & \checkmark & \checkmark & \checkmark & \checkmark \\ \hline
\end{tabularx}
\end{table*}

\section{Related Work}
\label{sec:2}
In this section, we review the related work on the framework of UAV rescue systems, MOO of MARL-based UAV computation offloading, and generative artificial intelligence (GAI)-assisted reinforcement learning.

\subsection{Framework of UAV rescue systems }
Extensive research has been done on the use of UAVs in emergency networks. Their high mobility, flexible deployment, and strong adaptability make them a key asset in supporting various applications. UAVs, for instance, can serve as aerial base stations, improving the connectivity of ground-based wireless systems. In \cite{tang2023disaster}, the authors formulated a mixed-integer nonlinear programming (MINLP) problem to maximize efficiency while optimizing UAV association, transmission power, and UAV location. In \cite{khalid2023computational}, the authors explored a UAV-assisted emergency communication system in post-disaster areas. The method mitigates the additional latency associated with long-distance data communication. The authors in \cite{wang2023mission} focused on maximizing the spatial exploration ratio while minimizing energy consumption and maintaining connectivity in post-disaster operations. The authors in \cite{sun2024joint} proposed a three-layer computing architecture integrating UAVs, MEC, and vehicle fog computing, along with a joint optimization problem for resource allocation.

Although these works focus on the navigation, communication, and civilian applications of UAVs, the frameworks ignore the feasibility of various types of robots in executing tasks, particularly from the perspectives of comprehensiveness and practicality in low-altitude UAV rescue.

\subsection{MOO of MARL-based for UAV Computation Offloading}
Task offloading in UAV networks using MARL is mainly divided into policy-based and value-based approaches. Examples of policy-based methods include MADDPG \cite{du2023maddpg}, multiagent twin delayed DDPG (MATD3)\cite{zhao2022multi}, and multiagent proximal policy optimization (MAPPO) \cite{liu2023energy}. The authors in \cite{zhao2022multi} proposed a cooperative multi-agent deep reinforcement learning framework to derive the joint strategy for trajectory design, task allocation, and power management. In \cite{du2023maddpg}, the authors proposed a framework for multiaccess MEC in air–ground networks and employed a MADDPG algorithm for efficient, adaptive decision-making under complex constraints. The authors \cite{liu2023energy} proposed a UAV-assisted MEC network with a digital twin to enhance service for mobile users. They formulated a resource scheduling problem as an MDP and emphasized the role of MAPPO in optimizing computation offloading. The value-based methods primarily include QMIX \cite{yu2024joint} and value-decomposition networks \cite{raivi2024jdaco}. The authors \cite{yu2024joint} proposed aerial edge computing networks using UAVs for computation offloading. A MARL algorithm based on QMIX was introduced to manage the complexity of joint computation offloading and trajectory optimization. The authors \cite{raivi2024jdaco} leveraged a MARL algorithm with value decomposition using a double deep Q-Network to optimize data aggregation and enhance offloading efficiency for UAV-enabled IoT systems in post-disaster contexts.

However, existing work has not comprehensively considered the practical challenges of computation offloading for low-altitude UAV rescue, such as obstacle constraints, resource availability in rescue operations, and system stability. To differentiate this work from existing works in the research area, a comparative analysis of various works is summarized in Table \ref{tab:table01}. 

\subsection{GAI-assisted Reinforcement Learning}
Generative AI is a prominent subfield of machine learning focused on the conceptualization and creation of content. It arises from the goal of enabling machines to generate novel and original data that accurately reflects the underlying patterns, structures, and nuances present in the training datasets. Generative AI encompasses a variety of advanced models, including generative adversarial networks (GANs), transformer-based models, and GDM, each employing distinct methodologies for learning and generating data \cite{zhang2024generative}. In \cite{li2024gan}, a GAN-driven auxiliary training mechanism for MARL is proposed. This approach reduces the overhead of real-world interactions and enables the agent's policy to be well-suited for real-world execution environments by performing offline training using generated environment status. The authors in \cite{chen2024transformer} combined Transformer with deep reinforcement learning (DRL) to address the scalability of the network and proposed a Transformer-based MARL algorithm for scalable multi-UAV area coverage. In \cite{du2024diffusion}, the authors integrated AI-generated optimal decisions with DRL to develop the deep diffusion soft actor-critic algorithm, improving the efficiency and effectiveness of selecting AI-generated content service providers. 

Although these works demonstrate that GAI holds substantial promise for enhancing the capabilities of multi-agent systems \cite{du2024diffusion}, they primarily focus on improving MARL's strategy generation while overlooking critical factors that influence this process, such as the observations of agents.

\section{System Model}
\label{sec:3}

In this section, we first introduce the application scenario and the proposed framework. We then define the mobility model, communication model, task completion latency model, and energy consumption model. The optimization problem is subsequently formulated.

\begin{figure}[!t]
	\centering
	\includegraphics[width=3.5in]{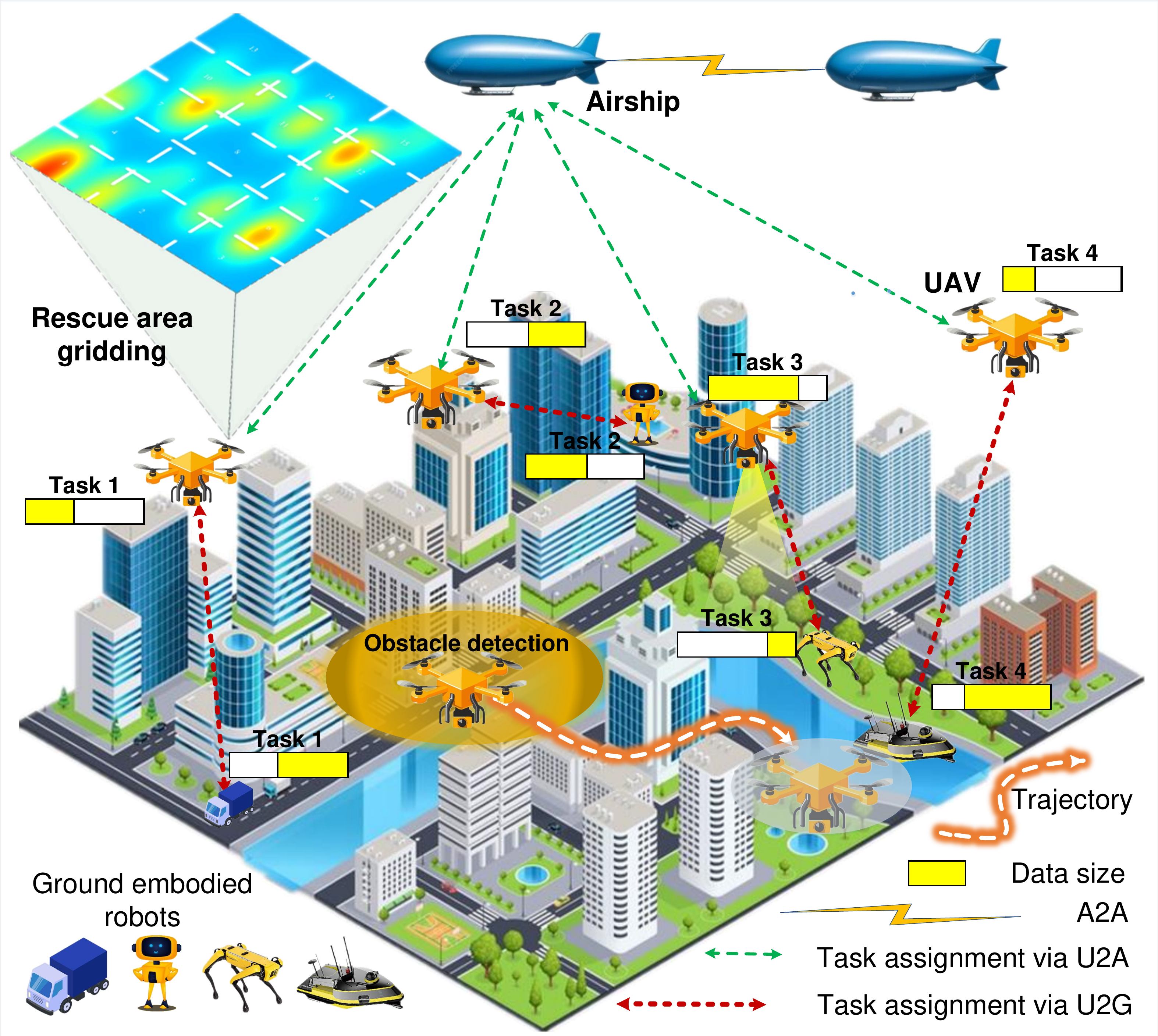}%
	\caption{System model. First, the GER computing power distribution map of the rescue area is divided into multiple subareas. Then, each UAV is assigned to multiple subareas. In each subarea, the UAV selects a suitable GER based on real-time observations to approach and perform computing task offloading.}
	\label{fig02}
\end{figure}

\subsection{Application Scenario and framework}
We consider a low-altitude UAV rescue mission in which multiple UAVs, operating at below 300 meters \footnote{ \url{https://www.scmp.com/topics/low altitude-economy}.}, are deployed to perform object detections using a convolutional neural network model, such as Yolov8s \cite{raivi2024jdaco, zhang2025robust}, in situations where the conventional communication infrastructure, including base stations, is unavailable \cite{raivi2024jdaco, zhang2025robust}. Fig. \ref{fig02} illustrates a system model framework for this application scenario. 
To improve the efficiency of rescue area exploration, UAV hovering is excluded from the scenario. Instead, UAVs process tasks online or offload them via U2G and U2A communications while in motion. Due to the disruption of the existing communication network, multiple airships are employed to provide communication coverage via airship-to-airship (A2A) communication. The airship, which has a longer flight duration and greater computational capabilities, hovers within 300-1000 meters higher than the UAVs, ensuring that all UAVs stay within its communication range \footnote{ \url{https://www.mskyeye.com/what-is-low altitude-economy/}.}. The UAVs within the communication range of the airship are represented by the set \( \mathcal{U} = \{1, 2, \dots, U\} \), and \( u \in \mathcal{U} \), where each UAV is assigned to multiple subareas, and these subareas do not overlap. Furthermore, we assume that both UAVs and GERs feature compatible communication interfaces, allowing for seamless networking and enabling UAVs to offload tasks to GERs. Let \( \mathcal{J} \), \( \mathcal{G} \), and \( \mathcal{B} \) represent the sets of GERs, airships, and subareas, respectively, where \( j \in \mathcal{J} \), \( g \in \mathcal{G} \), and \( b \in \mathcal{B} \). A three-dimensional Cartesian coordinate system \cite{42} is used to describe the locations of these entities. Without loss of generality, the UAVs operate at a constant hovering height $H$. The fixed hovering altitude minimizes energy consumption by avoiding frequent altitude changes due to terrain or buildings, which reduces the UAVs' movement energy \cite{43}. The parameters are presented in Table \ref{tab:table0111}.

\subsection{Mobility Model}
We divide the system timeline $T$ into $I$ discrete time intervals, $\mathcal{I} = \{1, 2, \ldots, I\}$ , and \( i \in \mathcal{I} \), each with equal length $\delta$, i.e., $T = I \delta$. Within the coverage area of an airship $g$, the numbers of UAVs and GERs are fixed for every time interval. Let $t_0$ signify the starting of $T$, and $t_i$ denote the time within the $i$-th interval, where $t_i \in [t_0 + (i-1)\delta, t_0 + i\delta]$. In the $i$-th time slot, the coordinate of UAV $u$ is given by 
$L_u(t_i) = \left[ x_u(t_i), y_u(t_i), H \right]$, where $[x_u(t_i), y_u(t_i)]$ is the horizontal coordinate of UAV $u$ in the $i$-th time slot. The starting and ending coordinates of UAV $u$ are predefined, denoted as $L_u^{start} = \begin{bmatrix} x_u^{start}, y_u^{start}, H \end{bmatrix}$ and $L_u^{end} = \begin{bmatrix} x_u^{end}, y_u^{end}, H \end{bmatrix}$, respectively. The coordinates of UAV $u$ in the $i$-th time slot remain fixed if $\delta$ is sufficiently small. By combining the coordinates of UAV $u$ across $i$ time slots, the Euclidean distance between the starting and ending coordinates of UAV $u$ is as follows:
\begin{equation}
\begin{aligned}
d_u(t_i) &= \| L_u(t_i) - L_u(t_{i-1}) \| \\
&= \sqrt{ \left( x_u(t_i) - x_u(t_{i-1}) \right)^2 + \left( y_u(t_i) - y_u(t_{i-1}) \right)^2}.
\end{aligned}
\end{equation}


The flight trajectory of UAV $u$ over time $T$ can be modeled as:
$L_u = \left\{ L_u^{start}, L_u(t_i), L_u^{end} \right\},$ where $L_u(t_i) = \left\{ L_u(t_1), L_u(t_2), \ldots, L_u(t_I) \right\}.$
The UAV's flight trajectory is modeled as a set of flight segments corresponding to each time slot. In the $i$-th time slot, the status of UAV $u$ is expressed as:
\begin{equation}
s_u(t_i) = \left( L_u(t_i), \theta_u(t_i), v_u(t_i) \right),
\end{equation}
where $\theta_u(t_i)$ is the angle between the UAV's flight tangent direction and the reference heading (i.e., the due north direction), and $v_u(t_i)$ is the real-time velocity of UAV $u$, which adheres to the maneuverability constraints. When UAV $u$ flies from the starting coordinate $L_u^{start}$ to the ending coordinate $L_u^{end}$, it can follow many optimal or suboptimal mobile trajectories. The UAV's trajectory planning can be defined as $s_u^{start} \overset{\mathfrak{S}}{\longrightarrow} s_u^{end}$, where $\mathfrak{S}$ denotes the set of all flight trajectories that meet the constraints.
\begin{table}[ht]
\caption{\textbf{Selected Symbols and Definitions}\label{tab:table0111}}
\centering
\begin{tabular}{c|c p{6cm}}
\hline
  &\textbf{Symbol}&   \textbf{Definition}\\
\hline
\multirow{21}{*}{\rotatebox{90}{\textbf{System Model}}}
 &$d_u(t_i)$ & Euclidean distance between the starting and ending coordinate of UAV $u$ in time slot $t_i$\\
 &$D_u(t_i)$ & Size of task \\
 &$l_{u,j}(t_i)$ & Channel power gain between UAV $u$ and GER $j$ at time slot $t_i$ \\
 &$L_u(t_i)$ & Coordinates of UAV $u$ at time slot $t_i$ \\
 &$s_u(t_i)$ & Status of UAV $u$ at time slot $t_i$\\
 &$E^{tota}(t_i)$ & Total energy consumption of a UAV\\
 &$\mathcal{I}$ & Set of time slots \\
 &$\mathcal{J}$ & Set of GERs\\
 &${M}$ & Set of computation resource allocations\\
 &$R_{u,j}(t_i)$ & Transmission rate between UAV $u$ and GER $j$ \\
 &$T_u^{UAV}(t_i)$ & The computing latency of UAV $u$ to process the task in time slot $t_i$ \\
 &$T_{u,j}^{GER}(t_i)$ & The total service latency of task processed by UAV $u$ at GER $j$ in time slot $t_i$\\
 &$T^{tota}(t_i)$ & Total task completion latency for all UAVs\\
 &$\mathcal{U}$ & Set of UAVs within the coverage of airship\\
 &${U}$ & Set of UAVs that choose to offload tasks to GERs\\
 &$\mathcal{G}$  & Set of airships \\
 &${W}$ & Set of UAV trajectory control decisions\\
 &${Z}$ & Set of risk sources\\
 &$\delta$ & Duration of each time slot \\
  \hline 
 \multirow{21}{*}{\rotatebox{90}{\textbf{Algorithm}}}&$A_i^u$ &  The agent's action of each time slot ${t}$\\
 &$L(Q(t_i))$ & The queue congestion \\
 &$\mathcal{L}_{t}$ & The loss function for the denoising network\\
 &$O_{i}$ & Set of agent observation\\
 &$p(\mathbf{x}_{t-1}|\mathbf{x}_t)$ &  The conditional probability distribution of the inverse process from time step $t$ to $t-1$\\
 &${Q(t)}$ & The virtual queue representing the accumulated energy consumption\\
 &$\mathrm{Q}$ &  The sets of critic networks for all agents\\
 &$\mathbf{x}_0$ & The observation of agent $n$ at the initial time\\
 &$\theta_{Q}$ & Sets of parameters for the critic networks \\
 &$\theta_{\pi}, \theta_{\pi^{\prime}}$  & Sets of parameters for the actor networks and target actor networks\\
 &$V$& The weighting factor for balancing task latency and energy consumption \\
\hline
\end{tabular}
\end{table}

\subsection{Communication Model}
\label{sec:3-3}
In low-altitude UAV rescue scenarios, both Line-of-Sight (LoS) and Non-Line-of-Sight (NLoS) conditions are considered for U2G. The channel power gain between a UAV and a GER is determined by integrating the probabilistic LoS transmissions with both small-scale and large-scale fading \cite{35}. For uplink communication, the channel power gain between UAV $u$ and GER $j$ during time slot $t_i$ can be expressed as:
\begin{equation}
\label{2}
l_{u,j}(t_i)=p_{u,j}^{L}l_{u,j}^{L}(t_i)+(1-p_{u,j}^{L})l_{u,j}^{NL}(t_i),    
\end{equation}
where $p_{u,j}^{L}$ represents the probability of LoS link, $l_{u,j}^{L}(t_i)$ and $l_{u,j}^{NL}(t_i)$ represent the gain between UAV $u$ and GER $j$ for LoS and NLoS links, respectively, which is defined as:
\begin{subequations}
    \begin{align}
l_{u,j}^{{L}}(t_i) = | \mathcal{H}_{u,j}^{{L}}(t_i) | ^{2}\left(\mathcal{L} _{u,j}^{{L}}(t_i) \right )^{- 1}10^{\frac {-\mathcal{F}_{\sigma }^{{L}}(t_i)}{10}},\label{3a} \\
l_{u,j}^{{NL}}(t_i) = | \mathcal{H}_{u,j}^{{NL}}(t_i) | ^{2}\left(\mathcal{L} _{u,j}^{{NL}}(t_i) \right)^{- 1}10^{\frac {-\mathcal{F}_{\sigma }^{{NL}}(t_i)}{10}}, \label{3b}   
    \end{align}
\end{subequations}
where $\mathcal{H}_{u,j}^{L}(t_i)$, $\mathcal{H}_{u,j}^{NL}(t_i)$, $\mathcal{L}_{u,j}^{L}(t_i)$, $\mathcal{L}_{u,j}^{NL}(t_i)$, $\mathcal{F}_\sigma^{L}(t_i)$ and $\mathcal{F}_\sigma^{NL}(t_i)$ are the components of path loss, shadowing, and small-scale fading for LoS and NLoS links, respectively. 

The small-scale fading of the channel is modeled using the Nakagami-$m$ fading model~\cite{36}. This model is parametric, scalable, and provides a good fit to the observed data. To avoid confusion with UAV identifiers, we use $w$ instead of $m$ to represent the shape parameter. Therefore, we refer to it as Nakagami-$w$. Specifically, $\mathcal{H}_{u,j}^{L}(t_i)$ and $\mathcal{H}_{u,j}^{NL}(t_i)$ follow the Nakagami distribution with fading parameters $w^{L}$ and $w^{NL}$, which can be given as:
\begin{subequations}    
\begin{align}
&\mathcal{H}_{u,j}^{{L}}(t_i)=\frac{2\left(w^{{L}}\right)^{w^{{L}}}h^{({2w^{{L}}-1})}e^{\left(-\frac{w^{{L}}h^{2}}{\overline{p}}\right)}}{\Gamma\left(w^{{L}}\right)\overline{p}^{w^{{L}}}},\label{4a}\\
&\mathcal{H}_{u,j}^{{NL}}(t_i)=\frac{2(w^{{NL}})^{w^{{NL}}}h^{({2w^{{NL}}-1})}e^{\left(-\frac{w^{{NL}}h^{2}}{\overline{p}}\right)}}{\Gamma\left(w^{{NL}}\right)\overline{p}^{w^{{NL}}}}\label{4b},
\end{align}    
\end{subequations}
where $\overline{p}$ represents the average power of the received signal in the fading envelope. $\Gamma(\cdot)$ represents the Gamma function. $h$ represents the signal amplitude.

The path loss between UAV $u$ and GER $j$ for LoS or NLoS link is defined as:
\begin{subequations}
\begin{align}
 \mathcal{L}_{u,j}^{{L}}(t_i)=\frac{\left(4\pi d_0f_c\right)^2}{c^2}\left(\frac{d_{u,j}(t_i)}{d_0}\right)^{\beta^{L}},\label{5a}\\
 \mathcal{L}_{u,j}^{{NL}}(t_i)=\frac{\left(4\pi d_0f_c\right)^2}{c^2}\left(\frac{d_{u,j}(t_i)}{d_0}\right)^{\beta^{NL}},\label{5b}   
\end{align}
\end{subequations}
where $f_c$ represents the carrier frequency, $c$ is the speed of light, $d_0$ is the reference distance, $d_{u,j}(t_i)$ is the distance between UAV $u$ and GER $j$, and $\beta^{L}$ and $\beta^{NL}$ are the path loss exponents for LoS and NLoS links, respectively.
Next, the shadowing refers to the large-scale signal attenuation caused by obstacles, and it can be modeled as a zero-mean Gaussian distributed random variable:
\begin{subequations}
\begin{align}
&\mathcal{F}_\sigma^{{L}}(t_i)\sim\mathcal{O}\left(0,\left(\sigma^{{L}}\right)^2\right),\label{6a}\\
&\mathcal{F}_\sigma^{{NL}}(t_i)\sim\mathcal{O}\left(0,\left(\sigma^{{NL}}\right)^2\right), \label{6b}
\end{align}  
\end{subequations}
where $\sigma^{L}$ and $\sigma^{NL}$ are the standard deviations of shadowing for LoS and NLoS links, respectively \cite{35}.

Accordingly, we use orthogonal frequency-division multiple access \cite{37} to reduce interference and support multiple UAVs simultaneously. In time slot $t_i$, the data transmission rate between UAV $u$ and GER $j$ can be given as:

\begin{equation}
\label{7}
R_{u,j}(t_i)={B_w}\log_2\:(1+\frac{P_ul_{u,j}(t_i)}{\sigma^2}),    
\end{equation}
where $B_w$ is the channel bandwidth and $P_u$ is the transmission power of the UAV. $\sigma^2$ is the noise power.

\subsection{Task Completion Latency Model}
\textbf{UAV computing latency.} The task of UAV $u$ generated in time slot $t_i$ is characterized as ${\{D_u(t_i),C_u,\tau_u}\}$, wherein $D_u(t_i)$ is the data size in bits, $C_u$ is the computation intensity of the task in cycles per bit, and $\tau_u$ denotes the deadline of the task. The service latency depends on the task offloading decision $\varsigma_{u,j}(t_i)$, which indicates the proportion or number of layers of the UAV's task offloaded to the GER. The computing latency of UAV $u$ to process task $D_u(t_i)$ locally in time slot $t_i$ can be given as:
\begin{equation}
\label{8}
T_u^{UAV}(t_i)=\frac{\varsigma_{u,j}(t_i)D_u(t_i)C_u}{f_u},    
\end{equation}
where $f_u$ denotes the computing capability of UAV $u$.

\textbf{GER computing latency.} The computing latency of UAV $u$ to offload task $D_u(t_i)$ to GER $j$ in time slot $t_i$ mainly consists of the transmission delay and processing delay. Specifically, the transmission delay can be given as:
\begin{equation}
\label{9}
T_{u,j}^{tran}(t_i)=\frac{(1-\varsigma_{u,j}(t_i))D_u(t_i)}{R_{u,j}(t_i)}.    
\end{equation}

Moreover, the processing delay can be expressed as:
\begin{equation}
\label{10}
T_j^{comp}(t_i)=\frac{(1-\varsigma_{u,j}(t_i))D_u(t_i)C_u}{f_{j,u}(t_i)},    
\end{equation}
where $f_{j,u}$ indicates the computing capability allocated by GER $j$ to the UAV $u$.

Given that the results are typically much smaller in comparison to the input data for most applications, the result download delay is ignored. Therefore, the service latency for GER computing latency and the total task completion latency for all UAVs can be calculated as:
\begin{equation}
\label{11}
T_{u,j}^{{GER}}(t_i)=T_{u,j}^{tran}(t_i)+T_j^{comp}(t_i),    
\end{equation}
\begin{equation}
\label{12}
\begin{aligned}T^{{tota}}(t_i)&=\sum_{u=1}^U\sum_{j=1}^J T_u^{UAV}(t_i)+T_{u,j}^{{GER}}(t_i).
\end{aligned}    
\end{equation}

\subsection{Energy Consumption Model}
\label{sec:3-5}
\textbf{Transmission energy consumption.}
The transmission energy consumption is approximated as the product of the transmission power $P_{u}$ and the transmission latency $T_{u,j}^{tran}(t_i)$ of the intermediate result of the task as follows:
\begin{equation}
\label{E08}
E_{u,j}^{tran}(t_i)= P_u T_{u,j}^{tran}(t_i).
\end{equation}

\textbf{Computation energy consumption.}
The computation energy consumption of UAV $u$ to process the task $K_u(t_i)$ in the slot $t_i$ can be given as:
\begin{equation}
E_u^{comp}(t_i)=\kappa_u(f_u)^2D_u(t_i)C_u,
\end{equation}
where $\kappa_u$ is the capacitance coefficient of UAV $u$, which is related to the chip structure of the CPU \cite{39}. 

\textbf{Propulsion energy consumption.}
The propulsion loss of UAV $u$ is dependent on its flight speed \cite{dai2023uav}. We assume that UAV $u$ travels at a constant speed $V_u(t_i)$ within each time slot, and the speeds can be different in different time slots. The propulsion energy consumption is given as follows:
\begin{equation}
E_{u}^{prop}(t_i)={w_u}v_u (t_i )^2,
\end{equation}
where $w_u$ is related to the weight of the UAV.

\textbf{Detection energy consumption.}
Throughout the entire flight of UAV $u$, there are unavoidable risks such as obstacles, adverse weather conditions, and complicated terrain. To estimate the energy expenditure for detecting risks during UAV's flight, we define the risk sources as a set $\mathcal{Z}=\{1, 2, \ldots, Z\}$, and \( z \in \mathcal{Z} \). In this case, the UAV's flight distance in the ${i}$-th time slot is divided into $m$ equal-length segments, and the Euclidean distance from the risk source $z$ to the central point of each segment is denoted by $\{d_{z_1}, d_{z_2}, \ldots, d_{z_m}\}$. When UAV $u$ travels from its starting coordinate $L_u^{start}$ to its ending coordinate $L_u^{end}$, its trajectory must be free of collisions. As such, UAV $u$ must monitor the specific rescue area of the risk source $z$ within each segment to ensure safe flying. Hence, the energy consumed by UAV $u$ in detecting the risk source $z$ during the ${i}$-th time slot is defined as $E_u^z(t_i)=\varpi\sum_{k=1}^m d_{z_k}$, where $\varpi$ is the unit energy consumed by UAV $u$ for sensing the risk source $z$ \cite{wang2022task}. When UAV $u$ encounters multiple risk sources from the set $\mathbb{Z}$ during flight, the total energy consumption for detection due to the surrounding risks can be expressed as:
\begin{equation}
E_u^{dete}(t_i)=\sum_{z\in\mathbb{Z}}E_u^z(t_i).
\end{equation}

To ensure that the UAV can provide long-term services at a low cost, the total energy consumption should be within an acceptable bound over a period of time, e.g., hours. Then, the total energy consumption is 
\begin{equation}
E^{tota}(t_i)=E_{u,j}^{tran}(t_i)+E_u^{comp}(t_i)+E_{u}^{prop}(t_i)+E_u^{dete}(t_i).
\end{equation}

We use $\bar{E}_{u}$ to denote the long-term average cost budget over time $T$. Then, a constraint is introduced to outline the system's long-term cost budget expectations for total energy consumption as follows \cite{dai2023uav}: 
\begin{equation}
\label{E1080}
\frac1T\sum_{i=1}^I\mathbb{E}[E^{tota}(t_i)]\leq\bar{E}_u.
\end{equation}


\subsection{Problem Formulation}
\label{sec:3:6}
The objectives of this paper are to reduce the overall task completion latency, minimize the total energy consumption of UAVs, and maximize the task offloading rate. To achieve these goals, three key elements are optimized: 1) The allocation of computational resources, denoted as $\mathbf{M} = \{m_{j,u}(t_i), \forall u \in \mathcal{U}, \forall j \in \mathcal{J}\}$; 2) The task offloading decisions, denoted as $\mathbf{F} = \{\varsigma_{u,j}(t_i), \forall u \in \mathcal{U}, j \in \mathcal{J}\}$, where $u$ represents a UAV that chooses to offload its task to a ground edge resource (GER); 3) The UAV trajectory control, denoted as $\mathbf{W} = \{w_j(t_i), \forall j \in \mathcal{J}\}$. Accordingly, the MOO problem is formulated as follows:

\allowdisplaybreaks

\begin{subequations}
    \begin{align}
        \mathbf{P1}:\min_{\mathbf{F, M, W}}& \{ \sum_{i=1}^I T^{tota}(t_i) \} \label{18} \tag{\theparentequation} \\
        \text{s.t.} \quad & \frac{1}{T}\sum_{i=1}^I \mathbb{E}[E^{tota}(t_i)] \leq \bar{E}_u, \forall u \in \mathcal{U}, \label{18a} \\
        & 0 \leq \varsigma_{u,j}(t_i) \leq 1, \forall u \in \mathcal{U}, j \in \mathcal{J}, \label{18b} \\
        & \sum_{j=1}^J \varsigma_{u,j}(t_i) \leq 1, \forall u \in \mathcal{U}, \label{18c} \\
        & \sum_{u=1}^U \varsigma_{u,j}(t_i) \leq 1, \forall j \in \mathcal{J}, \label{18d} \\
        & \varsigma_{u,j}(t_i) \cdot T^{UAV}(t_i) \leq \tau_u, \forall u \in \mathcal{U}, j \in \mathcal{J}, \label{18e} \\
        & \varsigma_{u,j}(t_i) \cdot T^{GER}(t_i) \leq \tau_{u}, \forall u \in \mathcal{U}, j \in \mathcal{J}, \label{18f} \\
        & 0 \leq f_{j,u}(t_i) \leq f_j^{\max}, \forall u \in \mathcal{U}, j \in \mathcal{J}, \label{18g} \\
        & \sum_{u=1}^{U} f_{j,u}(t_i) \leq f_j^{\max}, \forall j \in \mathcal{J}, \label{18h} \\
        & \delta \cdot d_u^{\min}/T \leq d_u(t_i) \leq \delta \cdot V_u^{\max}, \forall j \in \mathcal{J}, \label{18i} \\
        & \sum_{i=1}^I d_u(t_i) \leq L_u^{\max}, \forall u \in \mathcal{U}, j \in \mathcal{J}, \label{18j} \\
        & \| L_u(t_i) - L_{u+1}(t_i) \| \geq D, \forall u \in \mathcal{U}, j \in \mathcal{J}. \label{18k}
    \end{align}
\end{subequations}
where the constraint (\ref{18a}) represents the system’s long-term cost budget expectations for total energy consumption. $\bar{E}_{u}$ denotes the long-term average cost budget over time $T$. Constraints (\ref{18b})-(\ref{18d}) represent the task offloading constraints of UAVs. Constraints (\ref{18e}) and (\ref{18f}) indicate the deadline of the tasks. Constraints (\ref{18g}) and (\ref{18h}) limit the computation resource of UAV. Constraint (\ref{18i}) defines the mobility limitation of UAV $u$, where $V_u^{max}$ denotes the maximum velocity of UAV $u$. Constraint (\ref{18j}) specifies the flight distance of UAV $u$ during the $i$-th time slot. The flight trajectory of UAV $u$ consists of $I$ segments with varying lengths traversed during each time slot. $L_u^{max}$ indicates the maximum allowable flight distance. Constraint (\ref{18k}) addresses the spatial restriction for UAV $u$ and UAV $u+1$ operating in the $i$-th time slot, with $D$ representing the minimum safe distance between two UAVs. 

In general, the optimization problem outlined above cannot be solved in a single iteration due to the dynamic environment and the long-term objectives and constraints involved. Predicting dynamic channel conditions and user mobility over an extended period is highly challenging, and making real-time decisions under these long-term constraints is complex. Therefore, to efficiently solve \textbf{P1}, we employ an online algorithm capable of transforming the long-term optimization problem into an MDP. Problem \textbf{P1} involves integer variables (i.e., task offloading $\mathbf{F}$) and continuous variables (i.e., $\mathbf{M}$ and $\mathbf{W}$), while the inequalities in (\ref{18b})-(\ref{18j}) are non-convex constraints. Consequently, problem \textbf{P1} is an MINLP problem, which is also non-convex and NP-hard \cite{grossmann1997mixed}. Additionally, existing complexity analyses in \cite{meshkati2007energy} have formally confirmed this specific programming class as NP-hard.

\begin{figure}[!t]
    \centering
	\includegraphics[width=3.5in]{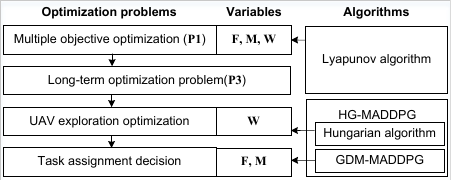}%
	\caption{The decomposition process of the optimization problem.}
	\label{fig03}
\end{figure}

As illustrated in Fig. \ref{fig03}, \textbf{P1} can be decoupled into a long-term deterministic optimization problem that incorporates stability considerations, which is addressed using the Lyapunov algorithm, denoted as \textbf{P3}. Additionally, the UAV exploration optimization problem is solved iteratively using the Hungarian algorithm. Subsequently, the task assignment decisions for the UAV, including computation resource allocation, task offloading ratio, and GER selection, are determined through the application of the HG-MADDPG method.


\section{Lyapunov-based Decoupling Method for Dynamic Long-term Problem}
\label{sec:4}
In this section, we first introduce the motivation for adopting Lyapunov optimization for the proposed problem and then use the Lyapunov algorithm to decouple problem \textbf{P1} into deterministic per-slot optimization problems.

\subsection{Motivation of adapting Lyapunov Optimization}
Lyapunov optimization is a powerful method for transforming long-term stochastic optimization into sequential per-slot deterministic problems while ensuring system stability \cite{liu2024dnn}. While  this paper focuses on an application scenario involving continuous and random tasks, where multiple objectives must be optimized simultaneously. The mobile edge computing terminals in this scenario include UAVs and GERs, which are deployed for rescue missions. Additionally, the stability of the system, composed of multiple resource-constrained mobile edge computing terminals, must be taken into account. We adapt the Lyapunov optimization method to the problem by using the Lyapunov function derived in Eq. \eqref{18csc} to jointly optimize energy consumption and task completion latency, and to quantify the accumulation of virtual queues $Q(t_i)$, thereby improving system stability. 


\subsection{Decoupling of \textbf{P1} Via Lyapunov Algorithm}
For problem \textbf{P1}, the key idea is to trade off the latency performance and the total energy consumption in the long run. In this section, the Lyapunov optimization method is employed to decouple the long-term optimization problem. Specifically, we introduce a virtual queue $Q(t_i)$ as the cost queue containing the accumulated energy consumption, as follows:
\begin{equation}
\begin{aligned}
\label{E008801}
 Q(t_{i+1})&=max\{Q(t_i)+E^{total}(t_i)-\bar{E}_u,0\}\\&=max\{Q(t_i)+y_k(t_i),0\},
 \end{aligned}
\end{equation}
where the value of $Q(t_i)$ represents the queue length, indicating the excess total energy consumption over the budget by the end of time slot $t_i$. To simplify the calculation, we add queue $E^{total}(t_i)$ and constant $\bar{E}_u$ to obtain $y_k(t_i)$. 

A large value of the virtual queue $Q(t_i)$ suggests that the current load status of energy consumption is likely to exceed the budget in the long run. Therefore, to satisfy the long-term constraint \eqref{18a}, the queue $Q(t_i)$ should be stable, as follows:


\begin{equation}
\label{18cc1}
 \lim_{T\to\infty}\frac{Q(T)}{T}=0.
\end{equation}

To further solve equation Eq. \eqref{18cc1}, we propose an online control algorithm for virtual queues based on the Lyapunov function, as outlined in Algorithm \ref{alg1}. This algorithm determines the optimal task offloading decision for the UAV in the current time slot. Next, the Lyapunov function is as follows:
\begin{equation}
\label{18csc}
L(Q(t_i))=\frac{1}{2}Q^2(t_i),
\end{equation}
where $L(Q(t_i))$ quantifies the congestion of the queue. The stability of the queue $Q(t_i)$ can be maintained if a policy function consistently drives the Lyapunov function towards a bounded value \cite{neely2010stochastic}. Next, we present a one-step conditional Lyapunov drift function $\Delta{L(Q(t_i))}$. The stability of $Q(t_i)$ can be attained by minimizing $\Delta{L(Q(t_i))}$ as:
\begin{equation}
\label{18cc}
\Delta{L(Q(t_i))}=\mathbb E\{L(Q(t_{i+1}))-L(Q(t_i))|Q(t_i)\},
\end{equation}

The drift-plus-penalty function of Lyapunov is expressed as:
\begin{equation}
\label{18bb}
\Delta L\left({Q}\left(t_i\right)\right)+V\mathbb{E}\left[\begin{aligned}T^{{tota}}(t_i)
\end{aligned} \right],
\end{equation}
where $V\geq0$ is a weighting factor to balance the total task completion latency and queue stability.

The problem \textbf{P1} can be converted into a series of deterministic problems for each time slot, given by 
\begin{subequations}
\allowdisplaybreaks[3]
\begin{align}
    \mathbf{P2}: \min_{\mathbf{F, M, W}} & \left\{ \Delta L\left(Q(t_i)\right) + V\mathbb{E}\left[T^{tota}(t_i)\right] \mid Q(t_i) \right\} \label{23} \tag{\theparentequation} \\
    \text{s.t.} \quad & 0 \leq \varsigma_{u,j}(t_i) \leq 1, \forall u \in \mathcal{U}, j \in \mathcal{J}, \label{23a} \\
    & \sum_{j=1}^J \varsigma_{u,j}(t_i) \leq 1, \forall u \in \mathcal{U}, \label{23b} \\
    & \sum_{u=1}^U \varsigma_{u,j}(t_i) \leq 1, \forall j \in \mathcal{J}, \label{23c} \\
    & \varsigma_{u,j}(t_i) \cdot T^{UAV}(t_i) \leq \tau_u, \forall u \in \mathcal{U}, j \in \mathcal{J}, \label{23d} \\
    & \varsigma_{u,j}(t_i) \cdot T^{GER}(t_i) \leq \tau_{u}, \forall u \in \mathcal{U}, j \in \mathcal{J}, \label{23e} \\
    & 0 \leq f_{j,u}(t_i) \leq f_j^{\max}, \forall u \in \mathcal{U}, j \in \mathcal{J}, \label{23f} \\
    & \sum_{u=1}^{U} f_{j,u}(t_i) \leq f_j^{\max}, \forall j \in \mathcal{J}, \label{23g} \\
    & \delta \cdot d_u^{\min}/T \leq d_u(t_i) \leq \delta \cdot V_u^{\max}, \forall j \in \mathcal{J}, \label{23h} \\
    & \sum_{i=1}^I d_u(t_i) \leq L_u^{\max}, \forall u \in \mathcal{U}, j \in \mathcal{J}, \label{23i} \\
    & \| L_u(t_i) - L_{u+1}(t_i) \| \geq D, \forall u \in \mathcal{U}, j \in \mathcal{J}, \label{23j} \\
    & \lim_{T \to \infty} \frac{Q(T)}{T} = 0. \label{23k}
\end{align}
\end{subequations}

Minimizing Lyapunov drift-plus-penalty function \eqref{18bb} needs the information of future time slots due to $\Delta L\left(Q\left(t_i\right)\right)$. To avoid involving future information, we derive and minimize its upper bound. By using the fundamental inequality, $\textit{max}\left\{x,0\right\}^{2}\leq x^{2}$, the upper bound of \eqref{18cc} is given by
\begin{equation}
\label{18ccss}
\begin{split}\Delta(Q(t_i))&=\mathbb{E}[L(Q(t_{i+1}))-L(Q(t_i))|Q(t_i)]\\&\leq\mathbb{E}[\frac12y_k(t_i)^2+Q(t_i)y_k(t_i))]\leq \Theta+Q(t_i)y_k(t_i),
\end{split}
\end{equation}
where $\Theta$ is a constant that upper bounds the first term on the right side of the above inequality. Such a constant exists because the $y_k(t_i)$ values are bounded \cite{neely2010stochastic}. 

A solution to problem \textbf{P2} can be obtained by minimizing the upper bound on the right-hand side of \eqref{23a} in each time slot, as given below:

\begin{subequations}
    \begin{align}
        \mathbf{P3}:\min_{\mathbf{F}} & \{ \Theta + V\mathbb{E}\left[T^{tota}(t_i)\right] + Q(t_i)y_k(t_i) \mid Q(t_i)\} \label{77} \tag{\theparentequation} \\
        \text{s.t.} \; & 0 \leq \varsigma_{u,j}(t_i) \leq 1, \forall u \in \mathcal{U}, j \in \mathcal{J}, \label{77a} \\
        & \sum_{j=1}^J \varsigma_{u,j}(t_i) \leq 1, \forall u \in \mathcal{U}, \label{77b} \\
        & \sum_{u=1}^U \varsigma_{u,j}(t_i) \leq 1, \forall j \in \mathcal{J}, \label{77c} \\
        & \varsigma_{u,j}(t_i) \cdot T^{UAV}(t_i) \leq \tau_u, \forall u \in \mathcal{U}, j \in \mathcal{J}, \label{77d} \\
        & \varsigma_{u,j}(t_i) \cdot T^{GER}(t_i) \leq \tau_{u}, \forall u \in \mathcal{U}, j \in \mathcal{J}, \label{77e} \\
        & 0 \leq f_{j,u}(t_i) \leq f_j^{\max}, \forall u \in \mathcal{U}, j \in \mathcal{J}, \label{77f} \\
        & \sum_{u=1}^{U} f_{j,u}(t_i) \leq f_j^{\max}, \forall j \in \mathcal{J}, \label{77g} \\
        & \delta \cdot d_u^{\min}/T \leq d_u(t_i) \leq \delta \cdot V_u^{\max}, \forall j \in \mathcal{J}, \label{77h} \\
        & \sum_{i=1}^I d_u(t_i) \leq L_u^{\max}, \forall u \in \mathcal{U}, j \in \mathcal{J}, \label{77i} \\
        & \| L_u(t_i) - L_{u+1}(t_i) \| \geq D, \forall u \in \mathcal{U}, j \in \mathcal{J}. \label{77j}
    \end{align}
\end{subequations}

It can be proved that Eq. \eqref{23k} always holds during the solution process of \textbf{P3}. Therefore, this constraint can be omitted when solving the problem \cite{neely2010stochastic}.
From \textbf{P3}, it can be obtained that solving the problem is equivalent to solving the MOO problem proposed in section \ref{sec:3:6}. By solving \textbf{P3}, the optimal solution of \textbf{P1} can be obtained, i.e., it can be obtained by
\begin{equation}
\label{E00880}
\varsigma_{u,j}\left(t_i\right)=\underset{{\varsigma_{u,j}\left(t_i\right)\in\mathbf{F}}}{\rm argmin}\mathbb{E}\left[\Theta+VT^{tota}(t_i)+Q\left(t_i\right)y_k\left(t_i\right)|Q(t_i)\right]. 
\end{equation}


$Remarks$. \textbf{P3} is freed from the constraint in Eq. \eqref{E1080} in comparison to \textbf{P1}. As a result, \textbf{P3} can be solved in an online fashion, without the need for global offline data. The objectives of \textbf{P3} are the task completion latency and the energy consumption of the UAV, which are weighted by a parameter $V$ and the energy consumption queue ${Q}({t_i})$, respectively. Adjusting these weighted factors allows for a balanced trade-off between task completion latency and the UAV's energy budget. Generally, $V$ provides a static adjustment that remains constant during GER-assisted UAV's task offloading. A higher $V$ helps in reducing the task completion latency. Additionally, the energy consumption queue ${Q}({t_i})$ provides dynamic control, which fluctuates based on the energy consumption of the UAV. An increased energy consumption queue indicates a lower remaining energy in the UAV, driving it to optimize energy usage in future time slots. This enables dynamic adjustment of the UAV's energy consumption in the optimization problem \textbf{P3}. Solving this online optimization yields efficient GER-assisted UAV task offloading for \textbf{P1}.

\begin{figure*}[htb]
    \centering
	\includegraphics[width=\textwidth]{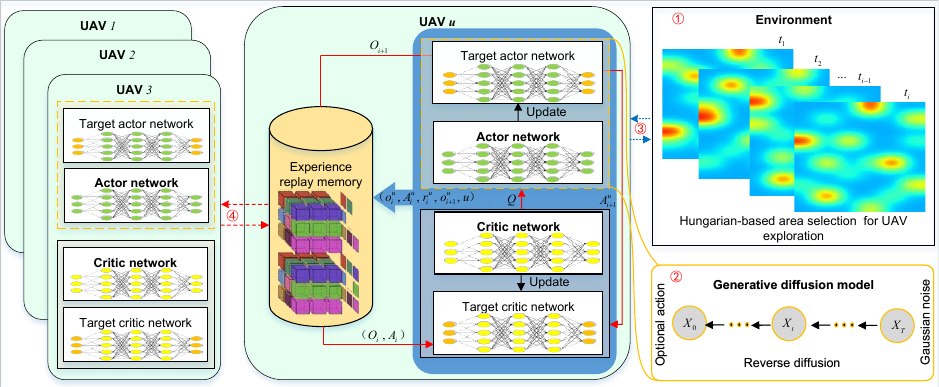}%
	\caption{Architecture of HG-MADDPG. \textcircled{1} Environmental observation, the HG-MADDPG makes each agent only needs to focus on the exploration subarea it selects, which reduces the dimension of the agent's observation space. \textcircled{2} Action generation, the HG-MADDPG designs the inverse diffusion process of GDM to replace the action network, which can generate the optimal decision in dynamic environment. \textcircled{3} Interaction between the agent and the environment, the agent based on the observations to realize the distributed execution to obtain their respective rewards. \textcircled{4} Agent-agent interaction, agents share experience by exchanging strategy sample sets, so that each agent has global training samples and realizes centralized training. }
	\label{fig04}
\end{figure*}

\begin{algorithm}
    \SetAlgoLined 
	\caption{ Lyapunov-Based Virtual Queue Online Control}\label{alg1}
	\KwIn{The observation of the UAV}
	\KwOut{The optimal task assignment decision in the current time slot $\varsigma_{u,j}\left(t_i\right) $}
        Initialize the queue $Q\left(0\right)=0$  
        
	\For {${i=1,2, \ldots,I}$}
{
		Update the queue by Eq. \eqref{E008801}
        
	Determine the optimal task assignment decision by Eq. \eqref{E00880}
    }
\end{algorithm}

\section{The Proposed HG-MADDPG Algorithm}
\label{sec:5}
In this section, we first provide an overview of the proposed HG-MADDPG algorithm. Next, we give our motivation for adopting the Hungarian algorithm for exploration area selection. Then, we introduce the motivation for adopting GDM and MADDPG, followed by elaborating on the offloading decision optimization for task assignment, modeling the problem as a MDP. We then present the interaction of the agent with the HG-MADDPG algorithm and conclude with an analysis of its computational complexity.

\subsection{Overview of the HG-MADDPG Algorithm}

The framework of the HG-MADDPG is shown in Fig. \ref{fig04}. The observation mechanism for the environment incorporates an area selection method based on the Hungarian algorithm, as illustrated in Algorithm~\ref{alg2}. This method eliminates the need for the agent (i.e., UAV) to acquire the status of all GERs or the task data size across different areas, thereby conserving the computational resources of the UAV. The reverse diffusion process seeks to recover the original data from noisy observations. Specifically, its training process iteratively predicts the noise distribution and trains the reverse diffusion model, enabling the agent to extract more information about data distributions from the environment's observations, generate actions, and execute them. The agent adjusts its parameters (such as denoising steps, batch size, and learning rate) based on feedback rewards, aiming to maximize long-term rewards and make optimal decisions. The task assignment and exploration optimization based on HG-MADDPG is detailed in Algorithm \ref{alg3}.

\subsection{Hungarian-based Area Selection for Exploration Optimization}
\label{sec:2222}
\subsubsection{Motivation of Adopting Hungarian Algorithm}
In low-altitude UAV systems with unknown environments, assigning flight areas effectively is crucial to maximize coverage and minimize redundancy or gaps. As the complexity of the environment increases, the observation space of agents expands, which introduces challenges for existing MARL methods like MADDPG. The motivation is driven by the need for optimal coverage while ensuring efficient UAV exploration. The Hungarian algorithm, with its optimal matching capability, ensures that each UAV is assigned to a specific area in a way that minimizes the total travel cost while covering all areas. Moreover, the Hungarian algorithm benefits from its low computational complexity, high efficiency, and stability, making it suitable for a variety of scenarios for solving matching problems \cite{cormen2009introduction}. Furthermore, the complexity is \( O(\max(U, \mathcal{B})^3) \), where \( U \) is the number of UAVs and \( \mathcal{B} \) is the number of subareas. This results in improved coverage efficiency, balanced workload distribution among UAVs, and reduced operational costs, making it a suitable choice for UAV exploration in unknown environments.

\subsubsection{Area Selection for Exploration Optimization}
The exploration optimization involves trajectory generation and obstacle detection (see Section \ref{sec:3-5}). Specifically, the UAV first selects an exploration area based on the Hungarian algorithm. Within the selected area, task offloading and GER selection are then performed using GDM-MADDPG. Once the GER is determined, the UAV’s trajectory is defined by its movement toward the selected GER to maximize the channel power gain (see Section \ref{sec:3-3}). During this process, the UAV also conducts low-altitude obstacle detection to avoid potential collisions. The Hungarian algorithm is based on the bipartite graph matching theory. It continuously searches for augmenting trajectories and updates matching relationships to achieve the optimal matching solution. In this paper, the predetermined rescue area is divided into \( \mathcal{B} \) subareas. UAVs incur varying costs when performing different tasks. The cost is as follows:
\begin{equation}
\label{E00990}
cost=\parallel L_{u}(t_{i})-L_{b}\parallel+D_{b}+C_{b}-e_{u}^{rema}(t_{i})-f_{b},
\end{equation}
where $L_b$ is the coordinate of the center point of the area, $e_u^{rema}$ is the remaining energy of the UAV, $D_b$ is the data size, $C_b$ is the area task calculation intensity (cycle/bit), and $f_b$ represents the average computing power of GER.

The cost matrix of each UAV to each area is represented by $\mathbb{C}[u][b]=cost_{u,b}$. The area selection approach based on the Hungarian algorithm is detailed in Algorithm \ref{alg2}. The specific steps are as follows:

\textbf{Step 1 (Line 1 of the Algorithm \ref{alg2}). } Initialize the label values of rows and columns, and the matching array.

\textbf{Step 2 (Line 2-line 9). } For each row (i.e., UAV) in the cost matrix, the algorithm finds the optimal matching area, i.e., the optimal matching area for each UAV. If the optimal matching solution cannot be found directly, the algorithm adjusts the label value to ensure that a new matching relationship can be found. This process continues to iterate until all UAVs are assigned to the corresponding areas.

\textbf{Step 3 (Line 10). } Based on the final matching relationship and the original cost matrix, the algorithm computes whether the total cost reaches the optimal area selection, i.e., the total cost is minimized.

\begin{algorithm}
    \SetAlgoLined 
	\caption{Hungarian-based Area Selection for UAV Exploration}\label{alg2}
	Initialize the label values and matching array: Set all the row labels $\alpha$ to the minimum value, i.e., $\alpha_{u}=\min\left(\mathbb{C}[u][1], \ldots,\mathbb{C}[u][\mathcal{B}]\right)$, set all the column labels $\beta_{b}=0$.\\
    \For {each UAV $u = 1, 2, \ldots, U$}
	{
        Search for a column that satisfies the matching condition $\mathbb{C}[u][b]\leq\alpha_u+\beta_b$ and has not been matched by any other row. Then match the current row with this column and mark the column as matched.

        Adjust the label values: \\
        During the process of finding an augmenting trajectory, adjust the row and column labels so that new matches can be found while maintaining the relationship between the cost matrix elements and the label values.

        \If {All areas are selected to UAVs}
        {
            Break the loop.
        }
	}

    Calculate the total cost.
\end{algorithm}

\subsection{HG-MADDPG-based Task Assignment and Exploration Optimization}
\label{sec:333}

\subsubsection{Motivation of Adopting GDM and MADDPG}
The motivation for adopting GDM and MADDPG lies in their ability to enhance decision optimization in multi-agent environments, especially with limited offline training data, and GDM's generative capabilities enable dynamic decision-making \cite{chen2024gainnet}. Moreover, integrating GDM and MADDPG into UAV systems refines training processes and optimizes decision strategies to enhance coordination and adaptability in unknown environments. Furthermore, the Hungarian algorithm and GDM are innovatively integrated into MARL, which respectively realizes the computational complexity of the observation space and the generation capability of the action space, thus improving the observation and execution effects of multiple agents.

\subsubsection{Markov Decision Process Modeling}
\label{sec:333-1}
The problem of exploration optimization and task assignment for UAVs is modeled as a multi-agent MDP, denoted by a tuple $(o_i^u, A_i^u, r_i^u, o_{i+1}^u, u)$, where $o_i^u$ represents the observation of the agent, $r_i^u$ indicates the reward received by the agent, and $o_{i+1}^u$ is the subsequent observation of the agent after performing the selected action. The agent learns an optimal task assignment strategy according to its own task number and computation resources, as well as the location and computing power of the GER, aiming to achieve the optimal task completion latency and energy consumption.

\textbf{Observations:} In each time slot ${t}$, the UAV collects environmental observations $O_i=\{o_i^1, o_i^2, \ldots, o_i^u\}$. Then, the observation of an agent is defined as:
\begin{equation}
\label{E0114}
o_i^u=
\begin{Bmatrix}
L_u(t_i), f_u, L_j, f_j
\end{Bmatrix},
\end{equation}
where $L_u$ and $f_u$ denote the UAV's coordinate and computing power, and $L_j$ and $f_j$ denote the GER's coordinate and computing power, respectively.

\textbf{Actions:} At time slot ${t}$, the action of an agent is $A_i^u$, which includes the task offloading target and the associated offloading ratio. The agent’s action is represented as:
\begin{equation}
\mathrm{A}_i^u = \{m_{u,j}(t_i), \mathrm{p}_{u,j}(t_i)\},
\end{equation}
where $m_{u,j}(t_i)$ represents the offloading GER of UAV $u$ at time slot $i$, and $\mathrm{p}_{u,j}(t_i)$ represents the offloading ratio at time slot $i$.

\textbf{Rewards:} Since the actions of agents are limited by energy consumption and directly impact task completion latency, the reward obtained for an action is given by
\begin{equation}
\label{E044}
r_i^u = \mathbb{E}[V \cdot T^{tota} + Q(t_i) y_k(t_i) | Q(t_i)].
\end{equation}


\subsubsection{UAVs Interacting with Environment}
\label{sec:333-2}
In a multi-agent setup, the agent's actor network generates specific actions based on its observations. These observations serve as inputs for the GDM's inverse diffusion process, which incrementally predicts and denoises the sampled Gaussian noise to generate actions according to the current conditions. For agent ${u}$, the purpose of the inverse diffusion process is to infer the task ratio $\mathbf{x}_0^u$ that is offloaded to the selected GER in the subarea from the Gaussian noise $\mathbf{x}_T^u{\sim}N(0,I)$. Since the number of GERs is random, the dimension of $\mathbf{x}_0^u$ depends on the maximum number of GERs in the subarea, that is, the dimension of the action space. If the probability $p(\mathbf{x}_{t-1}|\mathbf{x}_t)$ is learned, it is feasible to get sampling $\mathbf{x}_t$ from a standard normal distribution and samples from $p(\mathbf{x}_0)$ via the inverse denoising process. However, the estimation of $p(\mathbf{x}_{t-1}|\mathbf{x}_t)$ is computationally complex in practice. Therefore, the objective is to approximate $p(\mathbf{x}_{t-1}|\mathbf{x}_t)$ using $\boldsymbol{p}_\theta(\mathbf{x}_{t-1}|\mathbf{x}_t) = \mathcal{N}(\mathbf{x}_{t-1}; \boldsymbol{\mu}_\theta(\mathbf{x}_t,t), \boldsymbol{\Sigma}_\theta(\mathbf{x}_t,t))$. The probability from $\mathbf{x}_{T}$ to $\mathbf{x}_0$ can then be expressed as:
\begin{equation}
\label{E0088}
\boldsymbol{p}_\theta(\mathbf{x}_{0:T}) = \boldsymbol{p}_\theta(\mathbf{x}_T) \prod_{t=1}^T \boldsymbol{p}_\theta(\mathbf{x}_{t-1}|\mathbf{x}_t).
\end{equation}

The training of GDM involves optimizing the negative log-likelihood function of the training data. By adding conditional information $\mathbb{g}$, i.e., the agent's observation $o_i^u$, during the denoising process. The airship assigns subareas to achieve load balancing of UAV exploration. Therefore, the UAV only needs to handle the offloading decision of the assigned subarea to achieve the minimum task completion delay and energy consumption, which can not only maintain the optimal decision of a single subarea and the global optimal allocation but also reduce the computational complexity. At each time step, the conditioning the model predicts the parameters of the Gaussian distribution, specifically the mean $\mu_\theta(\mathbf{x}_t,t)$ and the covariance matrix $\boldsymbol{\Sigma}_\theta(\mathbf{x}_t,t)$. Based on \cite{ho2020denoising}, incorporating conditional information $\mathbb{g}$ during the denoising process enables the model to be treated as a prediction model for noise, and the covariance matrix is fixed to $\boldsymbol{p}_\theta(\mathbf{x}_{0:T})$ and $\Sigma_\theta(\mathbf{x}_t,\mathbb{g},t) = \beta_t {I}$. The mean is calculated as follows:
\begin{equation}
\label{E0039}
\mu_\theta(\mathbf{x}_t,\mathbb{g},t) = \frac{1}{\sqrt{\alpha_t}} \left(\mathbf{x}_t - \frac{\beta_t}{\sqrt{1-\overline{\alpha}_t}} \epsilon_\theta(\mathbf{x}_t, \mathbb{g}, t)\right).
\end{equation}

Next, $\mathbf{x}_T$ sampled from $N(0, I)$. The sampling dimension depends on the number of GERs in each subarea. Then, through the inverse denoising process parameterized by $\theta$, we sample $\mathbf{x}_{t-1} \mid \mathbf{x}_t$ as follows:
\begin{equation}
\mathbf{x}_{t-1} | \mathbf{x}_t = \frac{\mathbf{x}_t}{\sqrt{\alpha_t}} - \frac{\beta_t}{\sqrt{\alpha_t(1 - \overline{\alpha}_t)}} \boldsymbol{\epsilon}_\theta(\mathbf{x}_t, \mathbb{g}, t) + \sqrt{\beta_t} \boldsymbol{\epsilon},
\end{equation}
where $\boldsymbol{\epsilon} \sim N(0, I)$ represents a standard normal distribution, and $t = 1, \ldots, T$.

The loss function of the denoising network is defined by
\begin{equation}
\label{E01}
\mathscr{L}_t = \mathbb{E}_{\mathbf{x}_0,t,\boldsymbol{\epsilon}} \left[\|\boldsymbol{\epsilon} - \boldsymbol{\epsilon}_\theta(\sqrt{\overline{\alpha}_t}\mathbf{x}_0 + \sqrt{1-\overline{\alpha}_t}\boldsymbol{\epsilon}, t)\|^2 \right].
\end{equation}

\begin{algorithm}
	\SetAlgoLined 
	\caption{HG-MADDPG-based Task Assignment and Exploration Optimization}\label{alg3}
	The parameters of the actor-critic network and the inverse process of GDM
	
	\For{$\text{episode}=0\to \mathbb{E}$}{
		Initialize the agent environment
		
		\For{$i=1, 2, \ldots, \mathbb{I}$}{
			Get the area selection of each agent through \textbf{Algorithm \ref{alg2}}
				
			\For{$u=1, 2, \ldots, U$}{
						Agent $u$ obtains observation $o_{i}^{u}$ of the selected area $b$
						
				\For{$t=T, \ldots, 1, 0$}{
					Gaussian noise $\boldsymbol{\epsilon}$ is predicted and denoised, obtaining $\mathbf{x}_{0}^{u}$
				}
				An action selected by agent $u$ according to $o_{i}^{u}$
			}
				
				Execute action and obtain ${r}_{i}^{u}$ and the next $o_{i+1}^{u}$
				
				$o_i^u\leftarrow o_{i+1}^{u}$
				
				\eIf{the experience replay memory isn't full}{
					Store the training sample $(o_i^u,A_i^u,r_i^u,o_{i+1}^u,u)$ into the memory
				}{
					Update the memory

				\For{$u=1, 2, \ldots, U$}{
					Samples are taken from the memory
					$(o_i^u,A_i^u,r_i^u,o_{i+1}^u,u), \forall{i}=1, 2, \ldots, I$
					
					Update the actor network by Eq. \eqref{eq:loss} and the critic network by Eq. \eqref{eq:gradient}

				}
				Update the parameters of the target network according to $\psi$
				} 
		}
	}
\end{algorithm}

\subsubsection{UAVs Interacting with Each Other}
\label{sec:333-3}
The experience information $(o_i^u,A_i^u,r_i^u,o_{i+1}^u,u)$ generated by the agent in the process of interacting with the environment is stored in the experience replay memory. Then, multiple agents share the experience by exchanging sample sets. Furthermore, the agent uses the time difference error to assign priorities to the experience information so that important data is sampled more frequently, thereby improving learning efficiency and performance. The critic network assesses the effectiveness of the action produced by the actor network through rewards, i.e., the impact of the generated action on the long-term reward. When each agent calculates the forward propagation of the critic network, it splices the observations of all agents, including itself, into the observation vector $O_i=\{o_i^1,o_i^2, \ldots,o_i^U\}$, splices the actions of all agents into the action vector $A_i=\{A_i^1,A_i^2, \ldots,A_i^U\}$, and uses $(O_i,A_i)$ as the input of the critic network and outputs a one-dimensional $Q$ value, i.e., $Q_{\theta_Q^u}(O_i,A_i)$. In other words, the agent uses the information of all other agents in the environment to centrally train its own evaluation network. Next, the target actor network calculates the action $A_{i+1}$ taken by the agent in the next observation through the sample $O_{i+1}$ in the experience replay memory, and then constructs the MSE loss function of $Q_{\theta_Q^u}(O_i,A_i)$ and $Q_{\theta_Q^u}(O_{i+1},A_{i+1})$ with the time difference error and uses gradient descent to update the parameter $\theta_Q^u$. The loss function and gradient formula are as follows:
\begin{equation}
\begin{split}
    \min_{\theta_{Q}^{u}} \text{loss} 
    &= \min_{\theta_{Q}^{u}} \mathbb{E}_{(o_{i}^{u}, A_{i}^{u}, r_{i}^{u}, o_{i+1}^{u}, u) \sim D}\Big[ \Big( Q_{\theta_{Q}}(O_{i}, A_{i})\\
    &\quad \ - \big( r_{i}^{u} + \gamma Q_{\theta_{Q}}(O_{i+1}, A_{i+1}) \big) \Big)^2 \Big],
    \label{eq:loss}
\end{split}
\end{equation}

\begin{equation}
\begin{split}
    \nabla_{\theta_{Q}^{u}} J(\theta_{Q}^{u}) 
    &= \nabla_{\theta_{Q}^{u}} \mathbb{E}_{(o_{i}^{u}, A_{i}^{u}, r_{i}^{u}, o_{i+1}^{u}, u)\sim D} \Big[ \Big( Q_{\theta_{Q}}(O_{i}, A_{i}) \\
    &\quad - \big( r_{i}^{u} + \gamma Q_{\theta_{Q}}(O_{i+1}, A_{i+1}) \big) \Big)^2 \Big],
\end{split}
    \label{eq:gradient}
\end{equation}

\noindent where \( Q = \{ Q_1, Q_2, \dots, Q_u \} \) represents the sets of critic networks for all agents. Let \( \theta_Q = \{ \theta_Q^1, \theta_Q^2, \dots, \theta_Q^u \} \) denote the sets of parameters for the critic networks. 

\subsubsection{Complexity Analysis of HG-MADDPG Algorithm}
\label{sec:333-41}

In this section, we analyze the computational and space complexity of the proposed algorithm from the training and execution stages of the model, respectively.

$\bullet$ \textbf{Training stage}: \textbf{The computational complexity} is given by $\mathcal{O}(2|\theta_{\mu}| + 2|\theta_Q| + \mathbb{E}\delta|\theta_{\mu}| + \mathbb{E}\delta H\xi + U(2|\theta_{\mu'}| + 2|{\theta_{Q'}}|))$, which can be broken down as follows \cite{xie2025joint}.
The complexity of {\itshape Parameters initialization of actor-critic network} is $\mathcal{O}(2|\theta_{\mu}| + 2|\theta_Q|)$, where $|\theta_{\mu}|$ and $|\theta_Q|$ are the number of parameters of the actor-critic networks, respectively.
The complexity of {\itshape Observation-action pair sampling} is $\mathcal{O}(\mathbb{E}\delta|\theta_{\mu}|)$, where $\mathbb{E}$ is the total number of episodes.
The complexity of {\itshape Experience replay memory collection} is $\mathcal{O}(\mathbb{E}\delta H \xi)$, where $H$ denotes the computational cost of the Hungarian algorithm, and $\xi$ denotes the complexity of interacting with the environment.
The complexity of {\itshape Target actor-critic network update} is $\mathcal{O}(U(2|\theta_{\mu'}| + 2|{\theta_{Q'}}|))$. The target network parameters are updated $U$ times. 
\textbf{The space complexity} denotes $\mathcal{O}(2|\theta_{\mu}| + 2|\theta_Q| + \zeta(2|o_i^u| + |\mathrm{A}_i^u| + 2))$, where $\zeta$ means the size of the memory. $|o_i^u|$ and $|\mathrm{A}_i^u|$ represent the dimension sizes of the observation and the action spaces, respectively. The complexity includes both the parameters of the neural network and the memory storing the tuples $(o_ i^u,A_i^u,r_i^u,o_{i+1}^u,u)$. 

$\bullet$ \textbf{Execution stage}: \textbf{The computational complexity} is $\mathcal{O}(|\theta_{\mu}|)$, which is due to action inference of the actor network with the corresponding observation. Therefore, \textbf{The space complexity} is also $\mathcal{O}(|\theta_{\mu}|)$.

\section{Experiment Setup and Performance Evaluation}
\label{sec:6}
In this section, we introduce the experimental setup and evaluate the performance of HG-MADDPG in terms of convergence, stability, task completion latency, exploration capability, and resource visualization.
\subsection{Experiment Setup}
\label{sec:6-1}

For the first input layer in the actor network, the dimension of the observation space determines the number of neurons, while the third output layer has a number of neurons equal to the action dimension. The second hidden layer contains 256 neurons. For the critic network, the first input layer's neuron count depends on both the observation space dimension and the number of agents, with the hidden layer also containing 256 neurons. The learning rate of the actor-critic networks is \( \gamma = 10^{-4} \), and the size of the mini-batch is 512. The discount factor is \( \gamma_m = 0.9 \), with the exploration rate \( \epsilon_0 = 0.9 \) and the decay rate \( \beta = 10^{-4} \). The setup includes an airship at 600 meters altitude and nine UAVs at 50 meters, covering a 50×50 km² rescue zone. The airship is centrally located, with UAVs evenly distributed. All UAVs are within the airship's coverage and communicate with each other. Yolov8s is used as the convolutional neural network model for performance evaluation in the experiments. A complete summary of the parameters and their corresponding values is presented in Table \ref{tab:table2}. To assess its advantages, a set of benchmark algorithms is also evaluated.

\begin{itemize}
\item{\textbf{MADDPG} is a decentralized actor–centralized critic algorithm designed primarily for environments with continuous action spaces \cite{du2023maddpg}. The algorithm enables each agent to learn its own deterministic policy while leveraging centralized training to learn global information, which facilitates coordination among agents in partially observable settings. However, MADDPG can face challenges in training stability and scalability as the number of agents grows.}

\item{\textbf{MAPPO} is a policy-gradient-based method adapted from PPO, which uses clipped surrogate objectives to improve training stability and sample efficiency \cite{liu2023energy}. MAPPO supports centralized training with decentralized execution and works well in both discrete and continuous action spaces.}



\end{itemize}
\begin{table}[t]
	\caption{Experiment parameters \cite{tang2025dnn, sun2024joint}\label{tab:table2}}
	\centering
    	\begin{tabular}{p{1cm}|p{4.5cm}|p{2cm}}
		\hline
		Symbol & Definition & Value(Unit) \\
		\hline
    		$U$ & Number of UAVs & $[3,10]$ \\
            \hline
		$J$& Number of GERs  &  $[75,300]$ \\  \hline
          $\mathcal{B}$  &Number of subareas  & $[15,50]$ \\ 
          
          \hline
           $I$  &Number of rounds  & 5 \\ 
          
          \hline
	    $\mathbb{L}_{I}$ & Data size of task& $[12.5,125]$ GB \\  \hline
          $v_u^{max}$ & UAV propulsion speed & 30m/s  \\
            \hline
		$e_u^{max}$ &  Maximum UAV  power & $200$ Wh \\
            \hline
           $f_u$& UAV computing power &  5 TFPLOPs\\
        \hline
          $v_q$& Average processing rate of UAV &  12.5GB/min\\
        \hline
        $C_b$& Computational complexity &  $[200,500]$cycle/bit\\
            \hline	
		$P_{n}$ & Maximum transmission power  &$[10,100]$ mW \\
		\hline
	    $B_w$ & Maximum bandwidth  &$[1,10]$ Gbps \\
		\hline	
		$P_N$ & Noise power & -115 dBm 
     \\ \hline
	    $f_j$ & Computing power of GER &$[0,10]$ TFPLOPs \\  

		\hline
	\end{tabular}
\end{table}

\begin{figure}[t]
	\centering
	\includegraphics[width=3.5in]{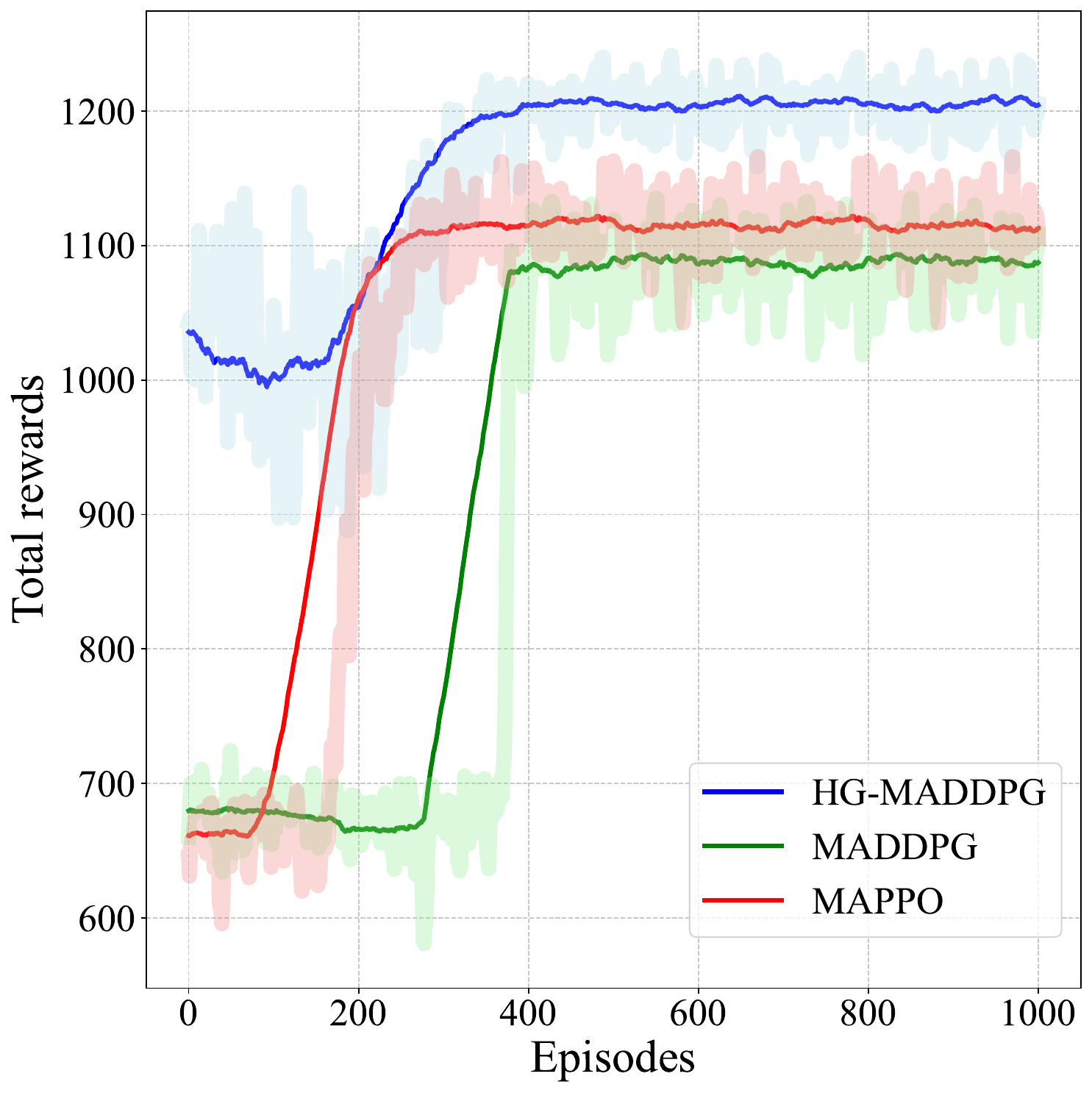}%
	\caption{Convergence comparison of different algorithms.}
	\label{fig05}
\end{figure}

\subsection{Performance Evaluation}

\subsubsection{Convergence of training Process}
\label{sec:6-2-1}

In Fig.~\ref{fig05}, although the convergence of the HG-MADDPG is slower, it consistently outperforms the MADDPG and MAPPO. The reward of HG-MADDPG increases steadily throughout the whole training process, eventually stabilizing at approximately 1200 after around 300 episodes. The improvement is due to the generative advantage of GDM, which enhances action sample efficiency by denoising steps.

\begin{figure*}[t!]
 	\centering
         \subfigure[Reward vs. denoising step.]{
 \includegraphics[width=0.31\textwidth]{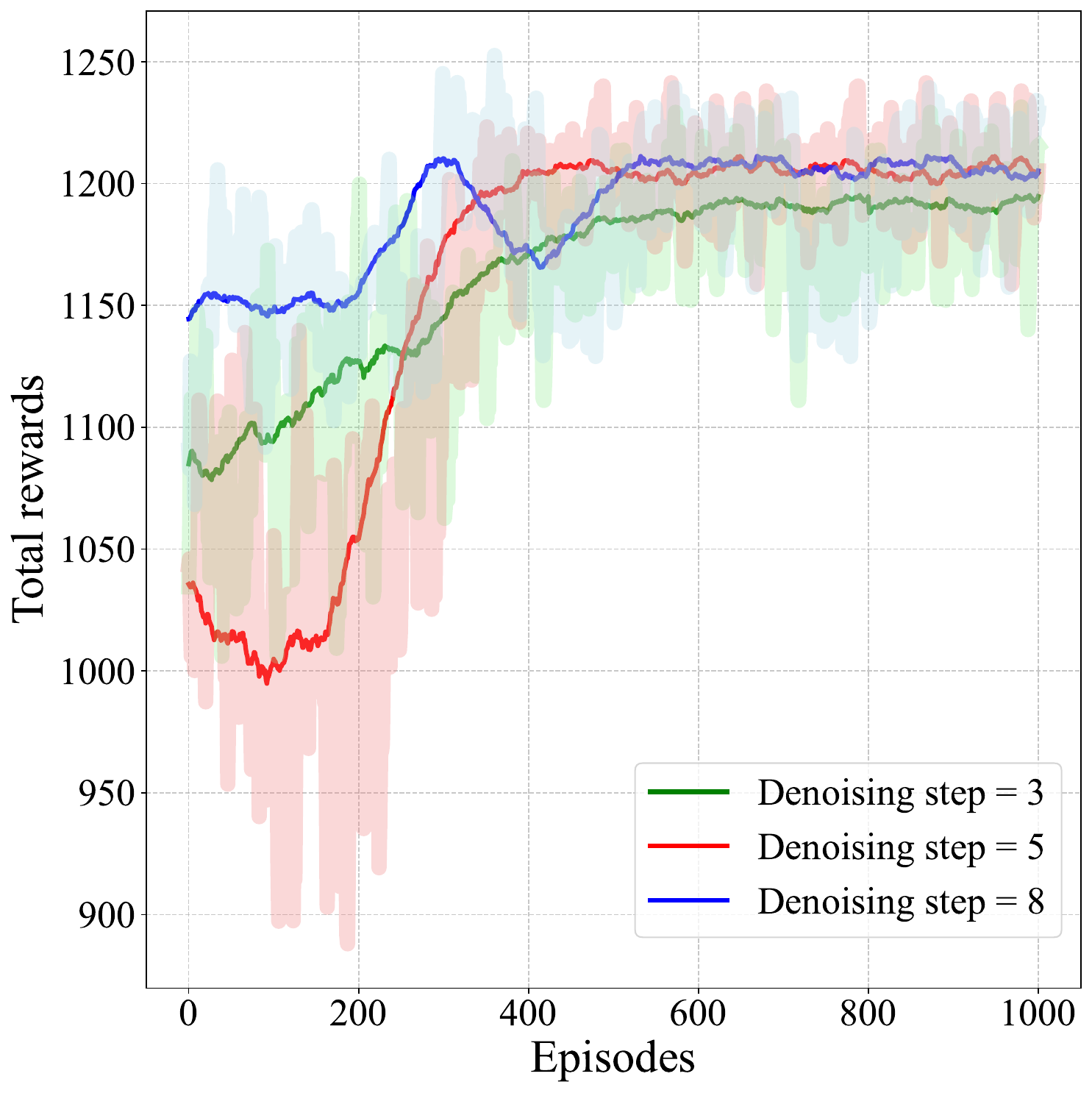}
 	}
        \hfill
         \subfigure[Reward vs. batch size.]{
 \includegraphics[width=0.31\textwidth]{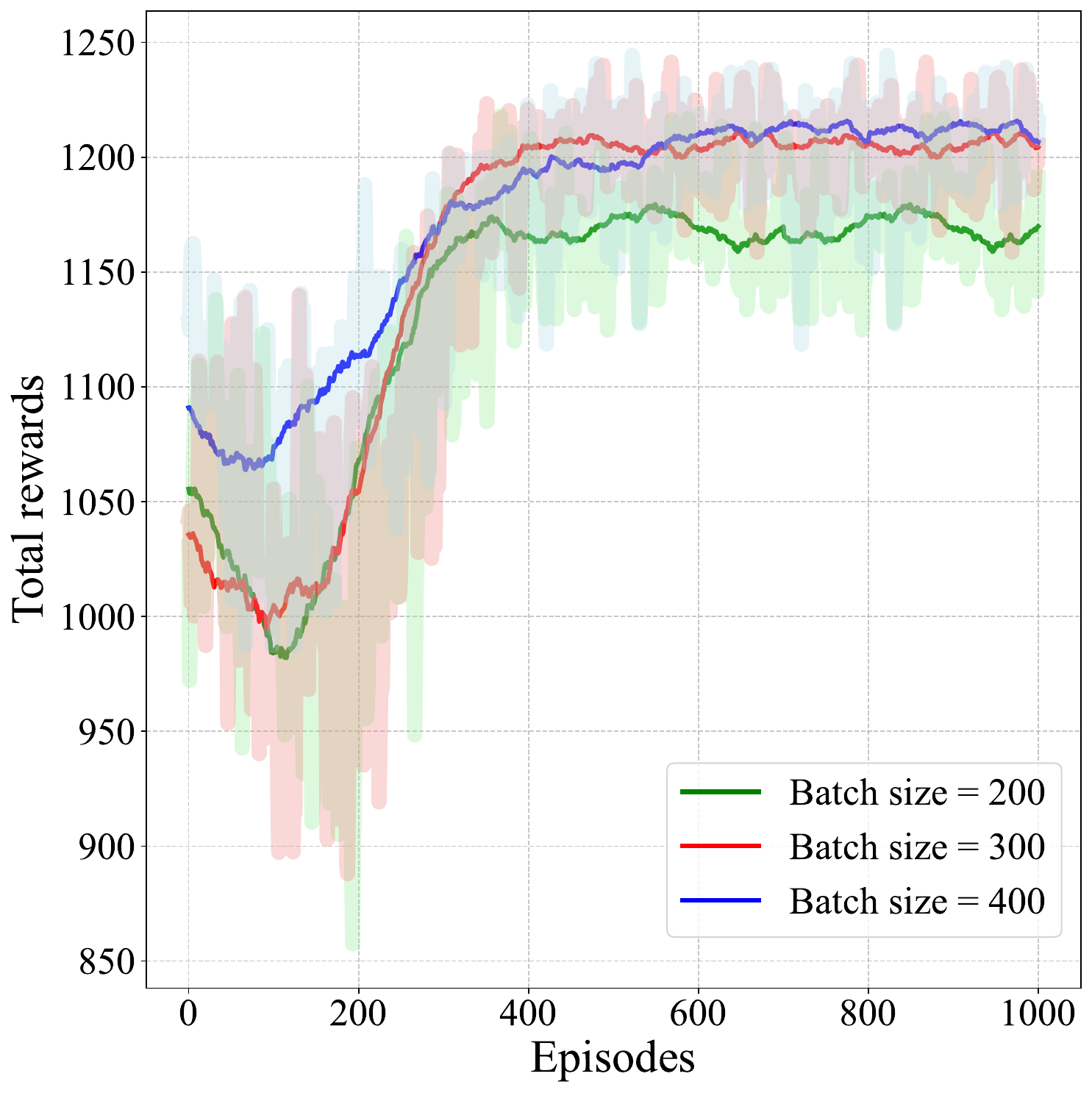}
 	}
        \hfill  
 	  \subfigure[Reward vs. learning rate.]{
 \includegraphics[width=0.31\textwidth]{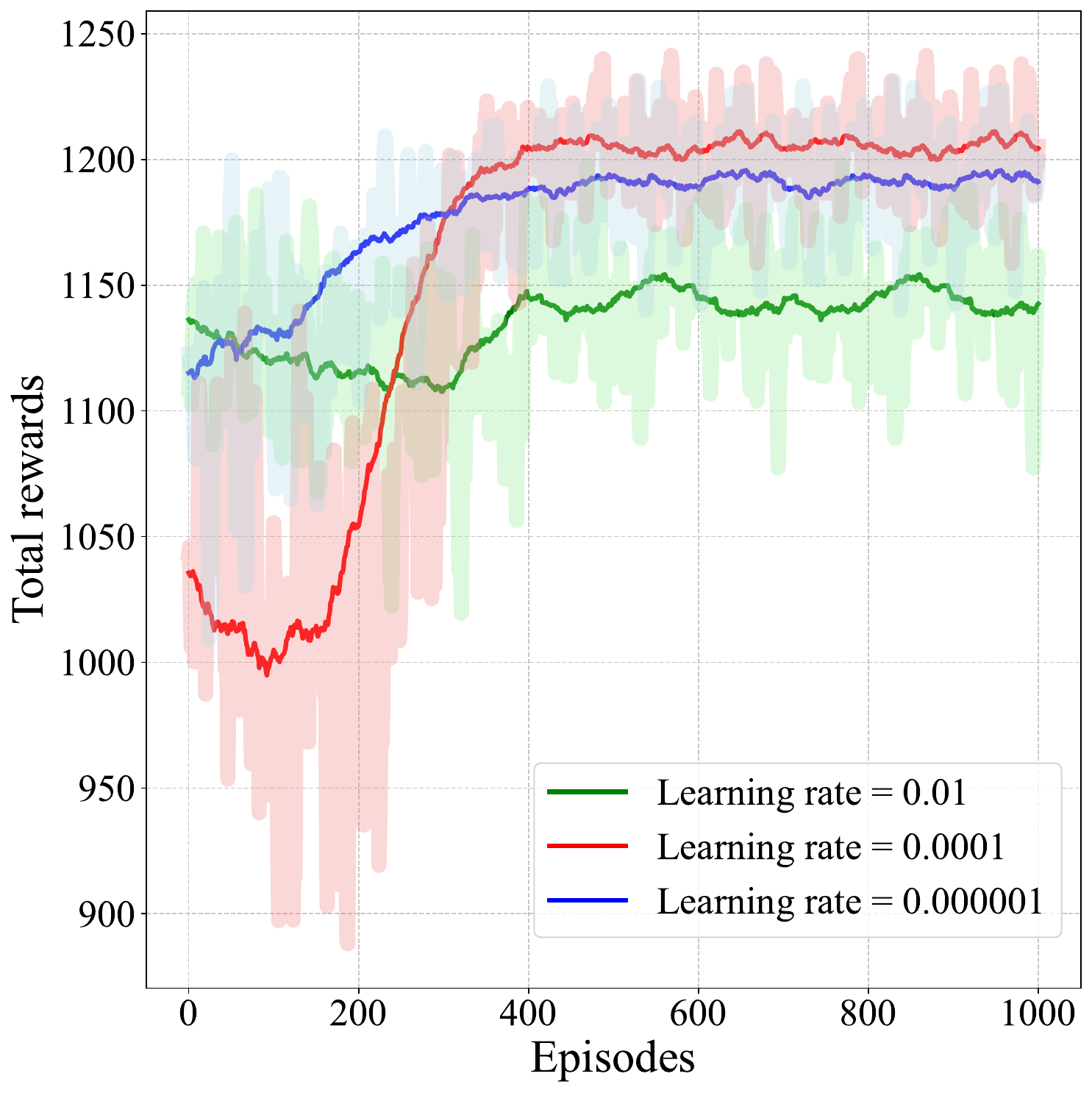}
 	}
 	\caption{The reward of different denoising steps, batch sizes, and learning rates.}
        \label{fig06}
 \end{figure*}

A set of experiments is performed to determine the optimal values for three key parameters affecting HG-MADDPG performance: denoising steps, batch size, and learning rate. From the results in Fig.~\ref{fig06}(a), it is evident that when the number of denoising steps is 5, the training process and the reward stability are superior to other step numbers. This suggests that five denoising steps are most effective for the denoising performance of the method proposed in this paper. Fig.~\ref{fig06}(b) shows that a batch size of 300 yields the best performance. Fig.~\ref{fig06}(c) demonstrates the convergence of the reward function under different learning rates. Clearly, the learning rates of 0.01 and 0.000001 do not support convergence, while a learning rate of 0.0001 accelerates convergence. Using a learning rate of 0.0001, we then evaluate the effect of batch sizes on the model training performance.

 \begin{figure*}[t!]
 	\centering
 	  \subfigure[Queuing energy vs. weighting factor.]{
 \includegraphics[width=0.31\textwidth]{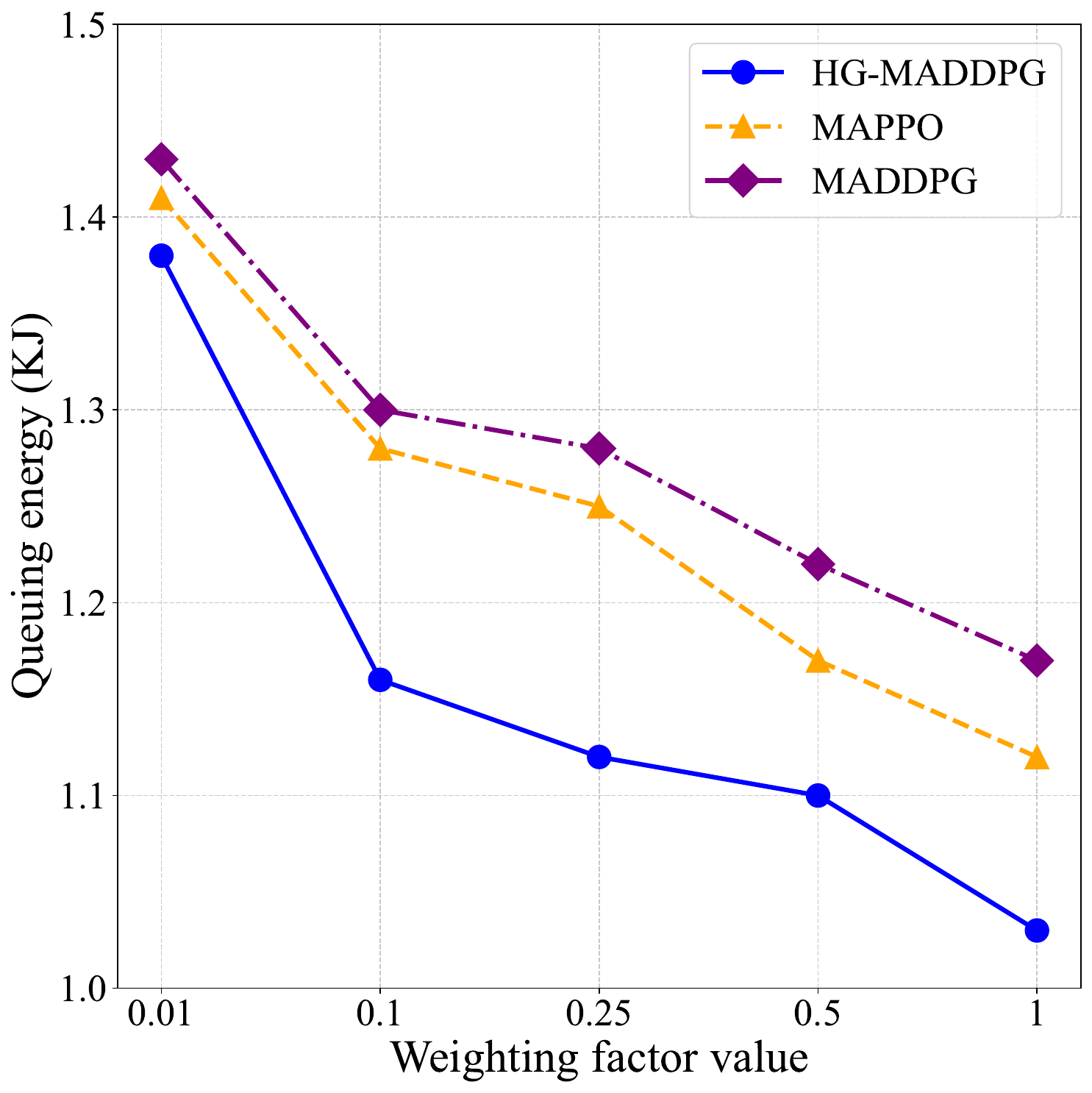}
 	}
        \hfill
         \subfigure[Queuing energy vs. computing power.]{
 \includegraphics[width=0.31\textwidth]{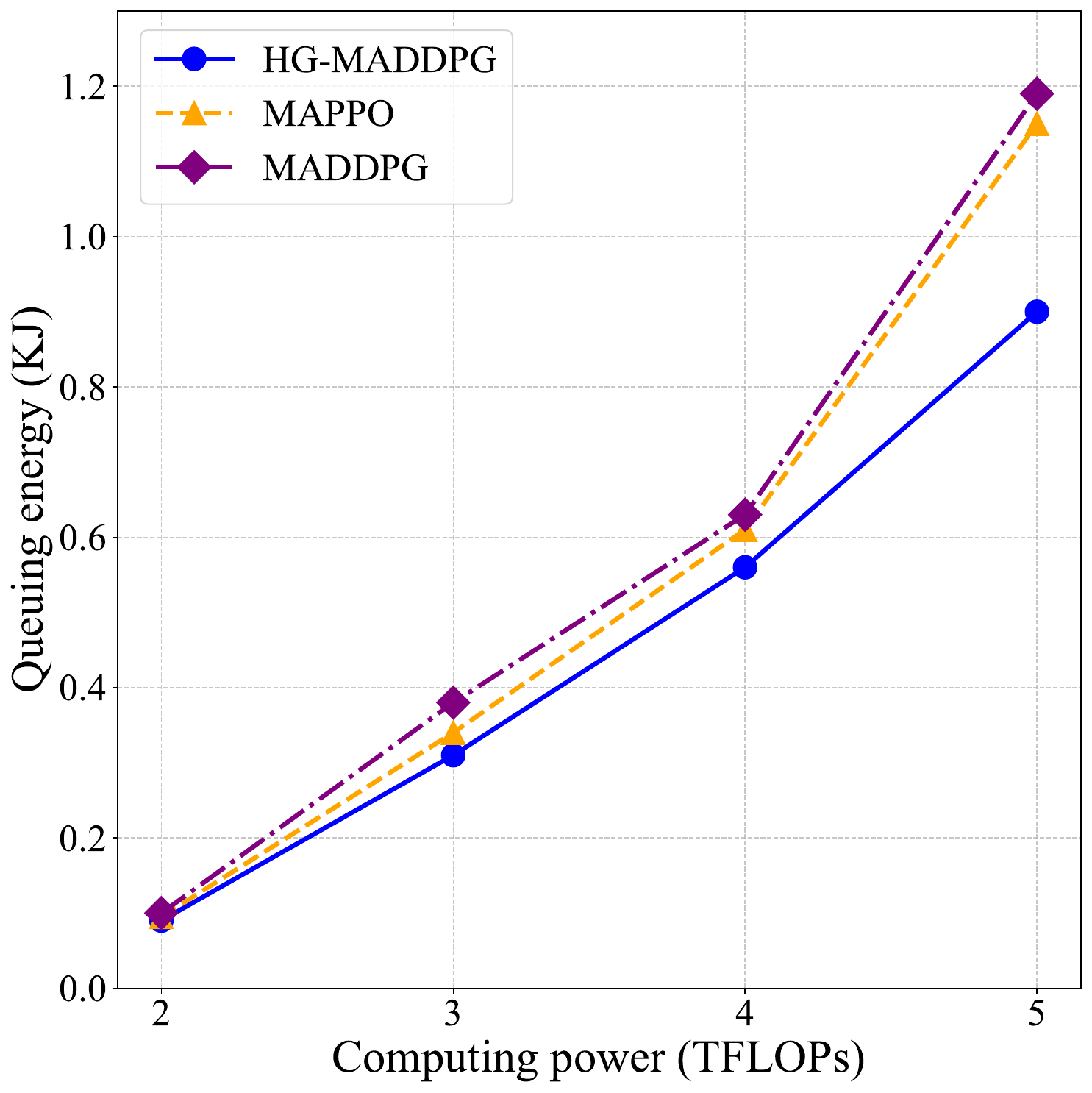}
 	}
        \hfill    
         \subfigure[Queuing energy vs. data size.]{
 \includegraphics[width=0.31\textwidth]{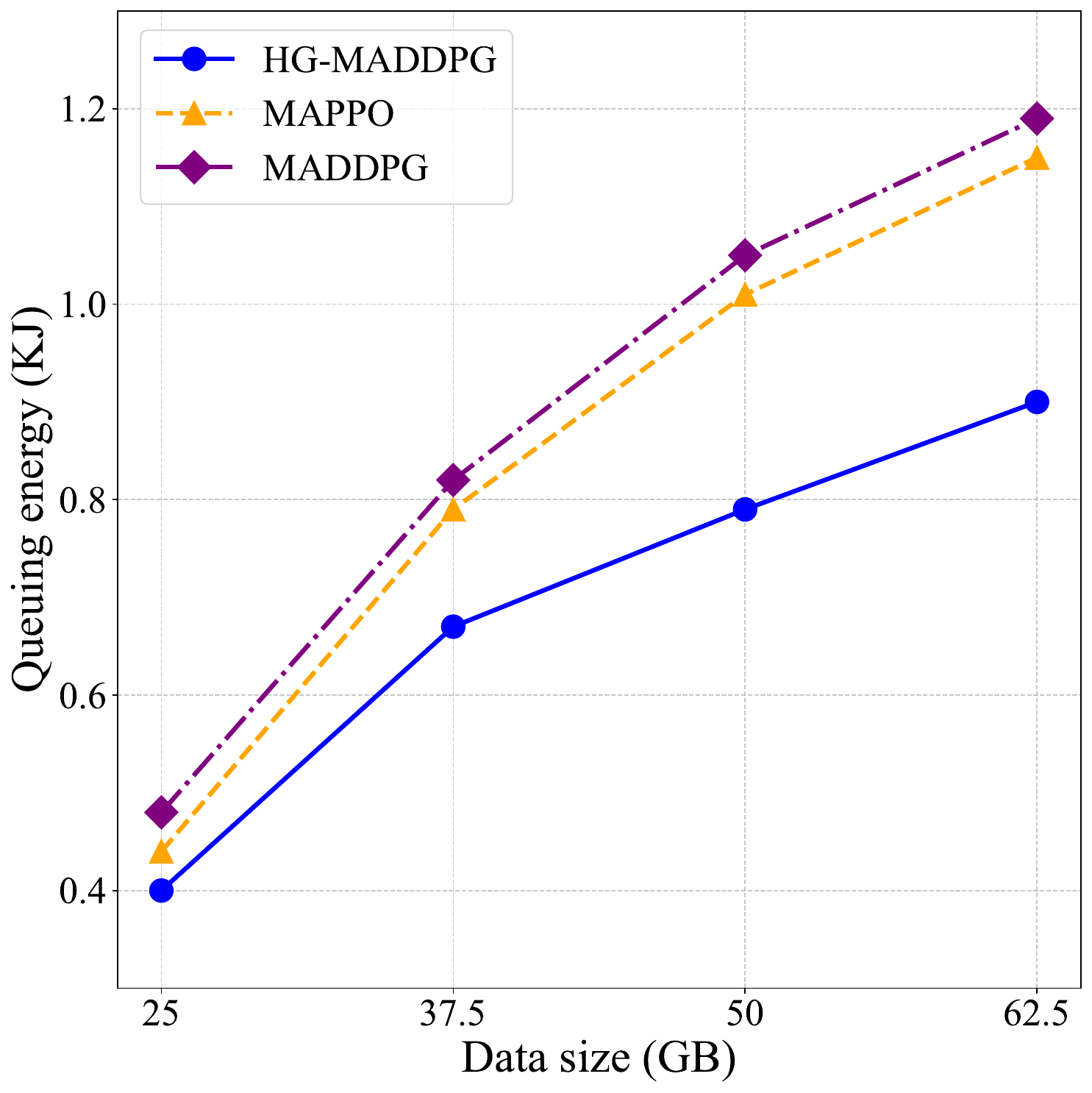}
 	}
 	\caption{Queuing energy of different weighting factor $V$, computing power, and data sizes.}
        \label{fig07}
 \end{figure*}

\begin{figure*}[t!]
 	\centering
         \subfigure[Latency vs. computing power.]{
 \includegraphics[width=0.31\textwidth]{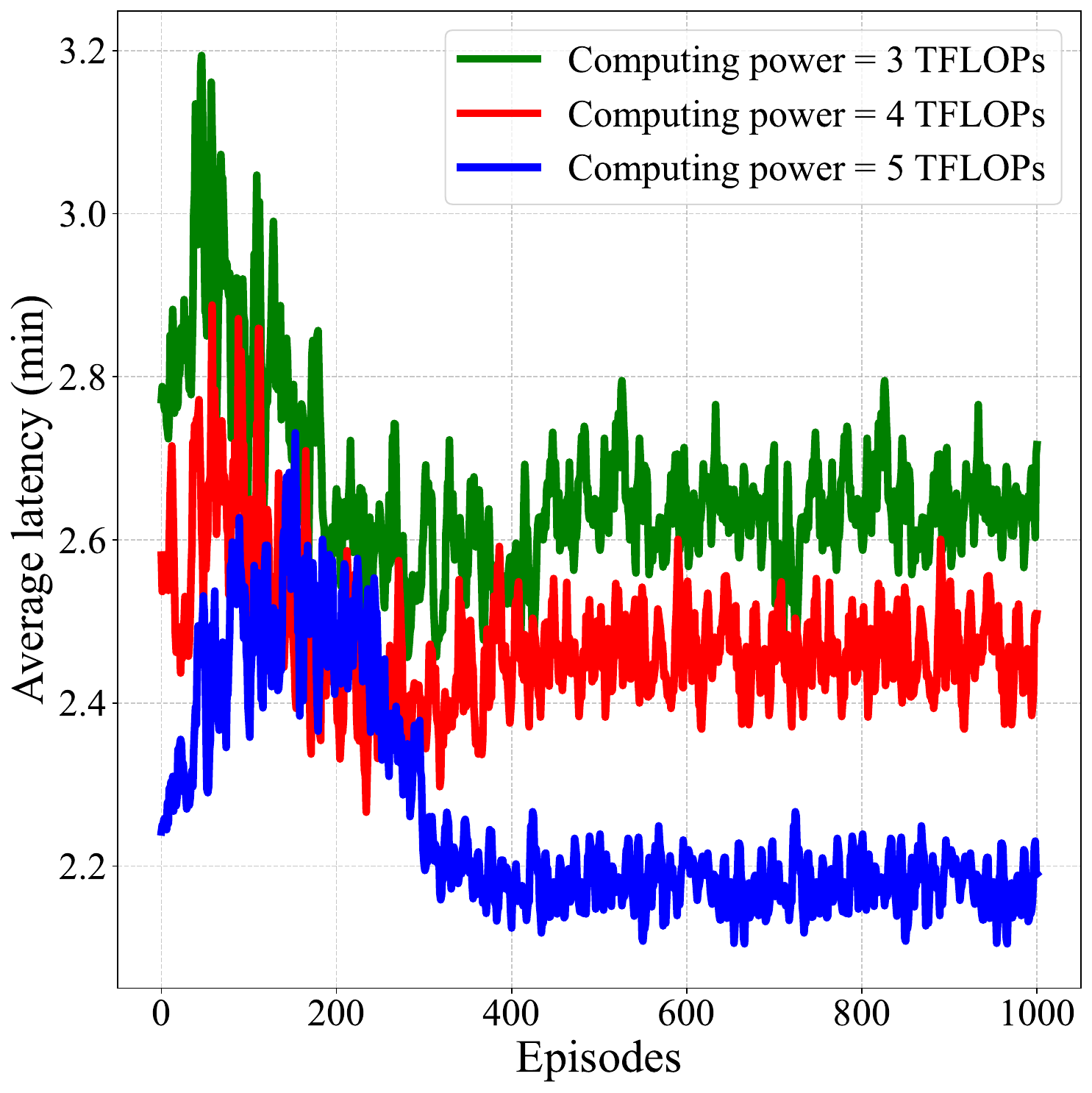}
 	}
        \hfill  
 	  \subfigure[Latency vs. number of GERs.]{
 \includegraphics[width=0.31\textwidth]{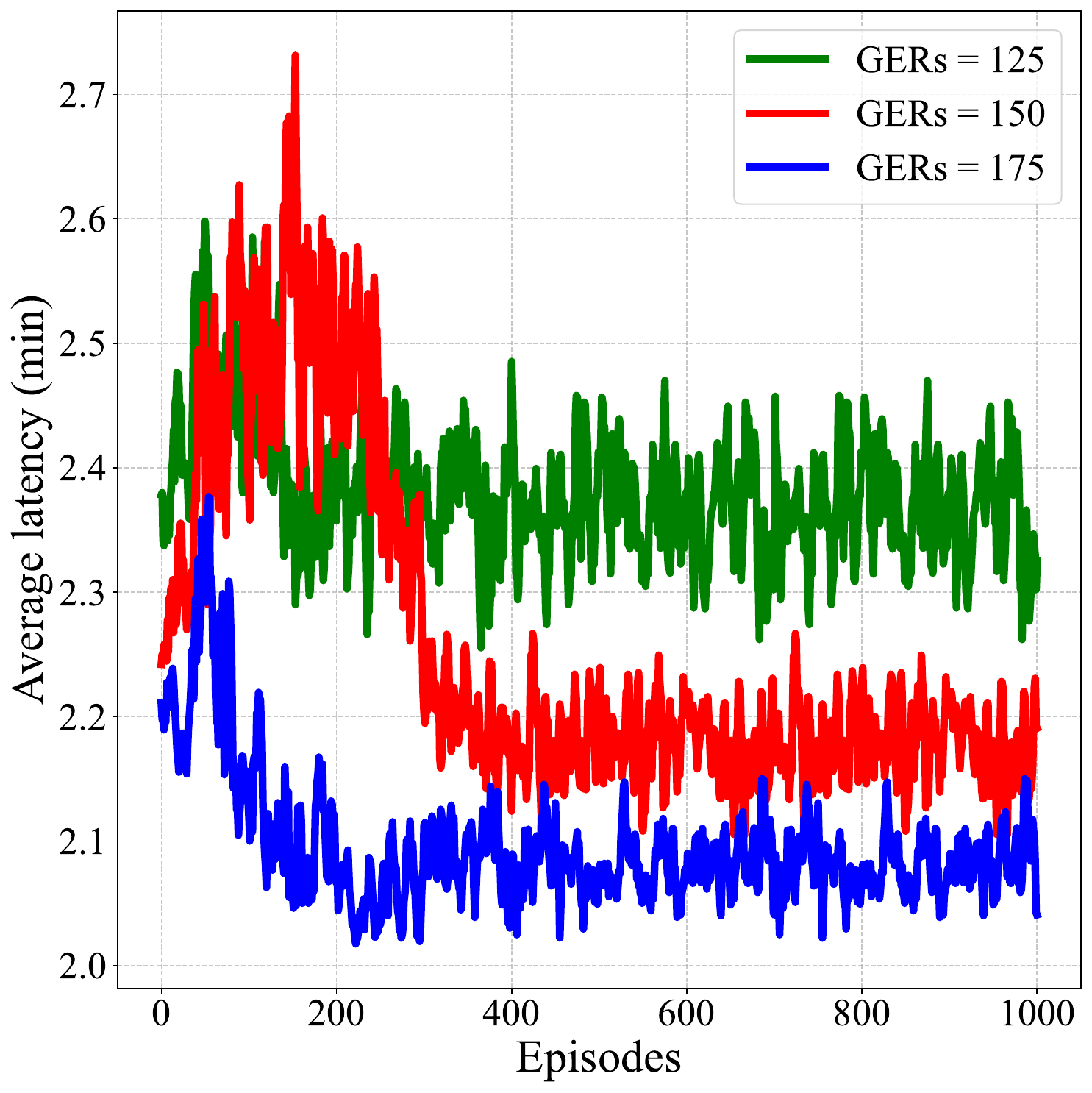}
 	}
        \hfill
         \subfigure[Latency vs. data size.]{
 \includegraphics[width=0.31\textwidth]{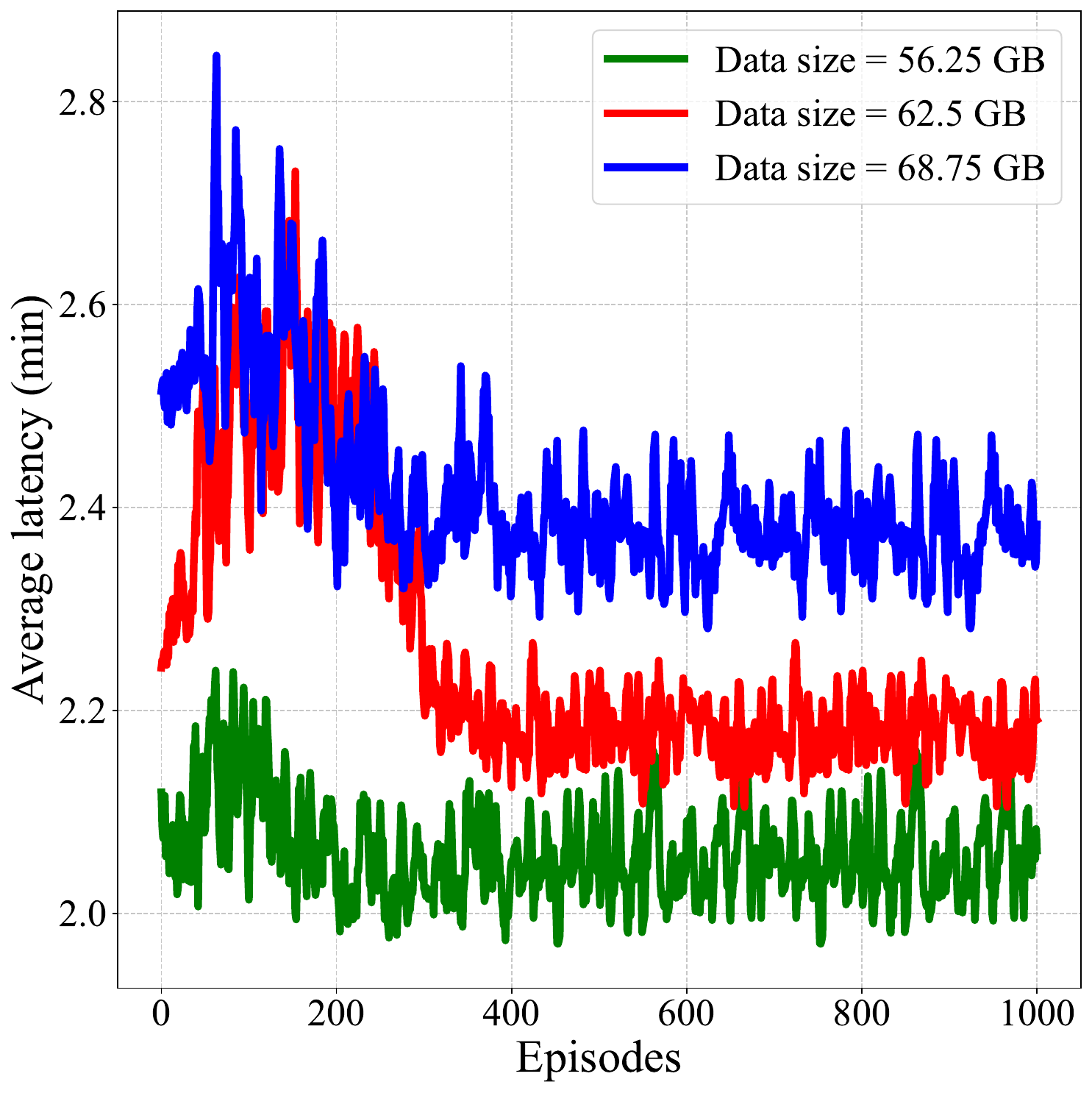}
 	}
 	\caption{Stability of task completion latency under different data sizes, computing powers, and number of GERs.}
        \label{fig08}
 \end{figure*}

\begin{figure*}[t!]
 	\centering
 	  \subfigure[Latency vs. computing power.]{
 \includegraphics[width=0.31\textwidth]{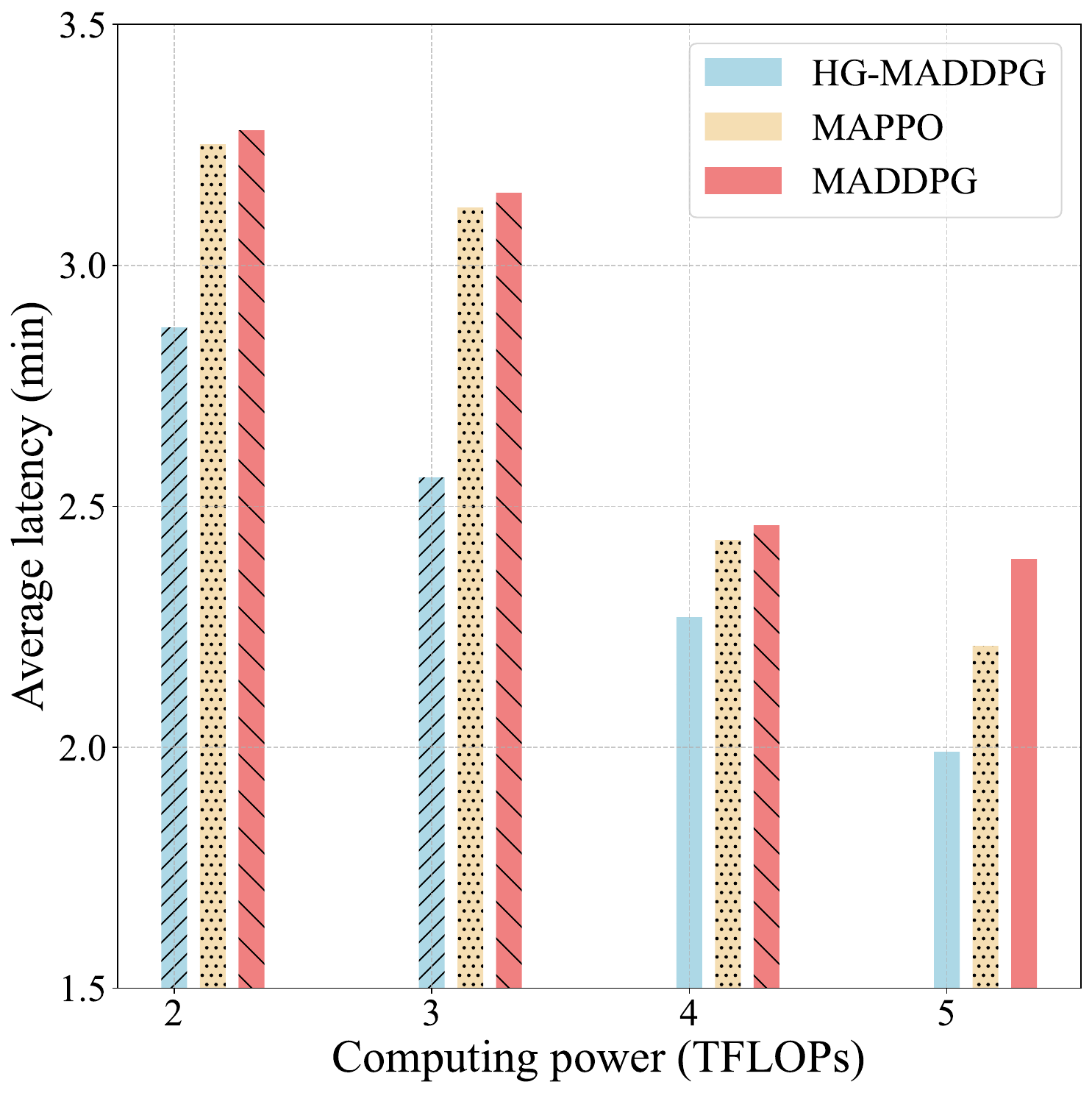}
 	}
        \hfill
         \subfigure[Latency vs. number of GERs.]{
 \includegraphics[width=0.31\textwidth]{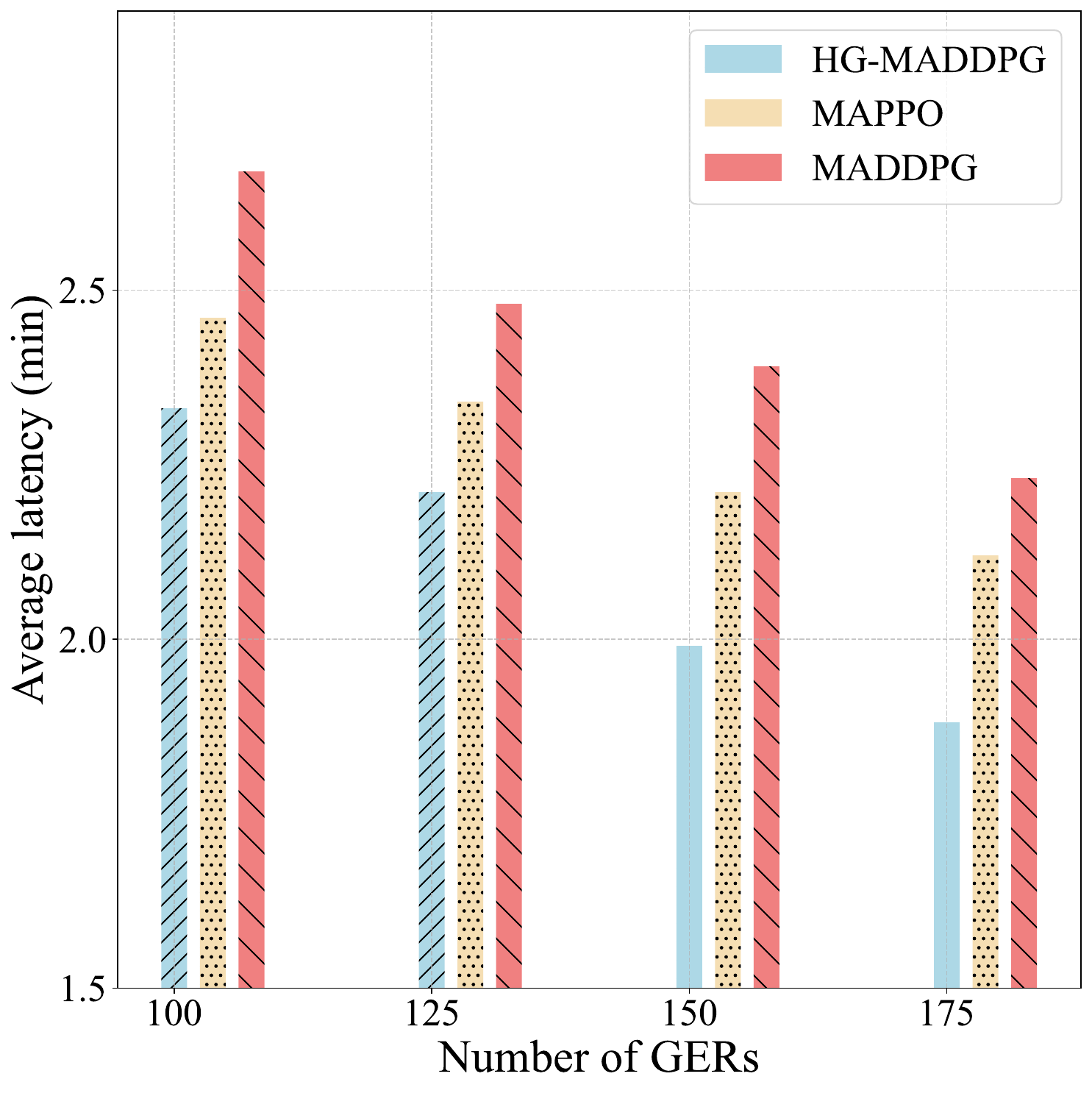}
 	}
        \hfill    
         \subfigure[Latency vs. data size.]{
 \includegraphics[width=0.31\textwidth]{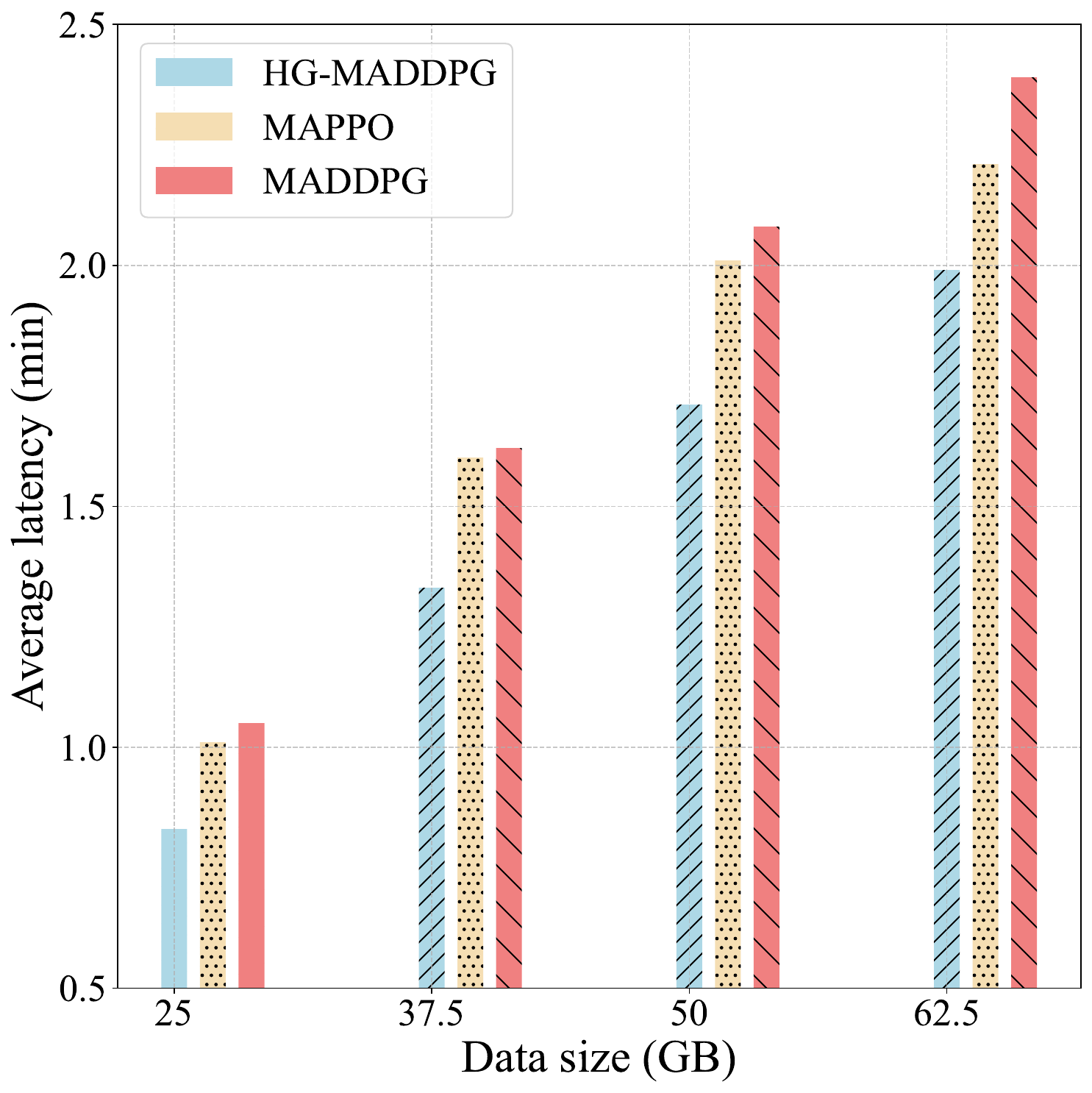}
 	}
 	\caption{Task completion latency of various algorithms under the condition of ensuring successful task completion.}
    \label{fig09}
 \end{figure*}
 
\subsubsection{Stability of Queuing Energy and Latency}
\label{sec:6-2-2}
In Fig.~\ref{fig07}, we investigate the queuing energy online control achieved by incorporating the Lyapunov technique into our approach. Fig.~\ref{fig07}(a) presents the queuing energy for various $V$ values. The queuing energy is monitored as the number of training episodes increases. Queuing energy refers to the average difference between the UAV's energy consumption in the current round and that in the previous episode. It reflects both the energy consumption trend across different task processing stages and the overall energy level of the task queue. The reason is that the diffusion model in HG-MADDPG is used to enhance the learning ability of the policy network so that it can more accurately understand the dynamic changes in energy consumption. Similarly, Figs.~\ref{fig07}(b) and \ref{fig07}(c) present the queuing energy under varying computing power and data sizes, respectively.

To further demonstrate the significant impact of the Lyapunov technique introduced in this paper on system stability, we focus on the stability requirements of task completion latency in low-altitude UAV rescue emergency systems. Verification experiments were conducted to assess the system's task completion latency under varying conditions of GER computing power, available computation resources, and data sizes. The task completion latency of the system remains within a relatively stable range, as illustrated in Fig. \ref{fig08}. The observed convergence behavior of these evolving task queues confirms the stability of task assignment, ensuring the reliable operation of the system as outlined in this study. 

\subsubsection{Task Completion Latency}
\label{sec:6-2-3}

The Fig.~\ref{fig09}(a) and (b) illustrate how the task completion latency is affected by progressively increasing the computing power and number of GERs. This experiment was conducted with 3 UAVs and \(V = 0.5\). As the computing power and number of GERs increase, the task completion latency decreases. HG-MADDPG achieves the largest reduction in average task completion latency, outperforming MADDPG and MAPPO by 20.35$\%$ and 12.56$\%$, respectively. This is because HG-MADDPG has a faster decision generation capability and has fewer observation space dimensions. Furthermore, as shown in Fig.~\ref{fig09}(c), while the computing power and number of GERs remain unchanged, as the task data size increases, the task queue becomes longer and the completion time increases accordingly. Compared with the other two baseline methods, the task completion time based on HG-MADDPG is the shortest. This is because the UAV based on HG-MADDPG will consider the current computing power of GER when selecting the offloading object and select the GER with strong current computing power and low task completion delay for task offloading. This shows that the above experimental results can clearly illustrate that the proposed method is feasible and in line with common sense.

\subsubsection{Altitude Exploration and Resource Visualization}
\label{sec:6-2-4}
\begin{figure*}[t!]
    \centering
    \subfigure[Subarea (colors) and 2D trajectory.]{\includegraphics[width=0.27\textwidth]{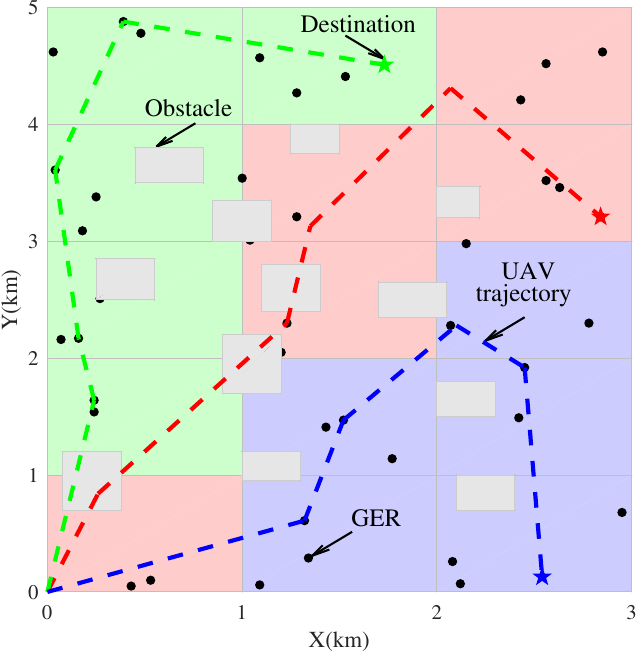}}
    \hfill
    \subfigure[3D trajectory and obstacle avoidance.]{\includegraphics[width=0.31\textwidth]{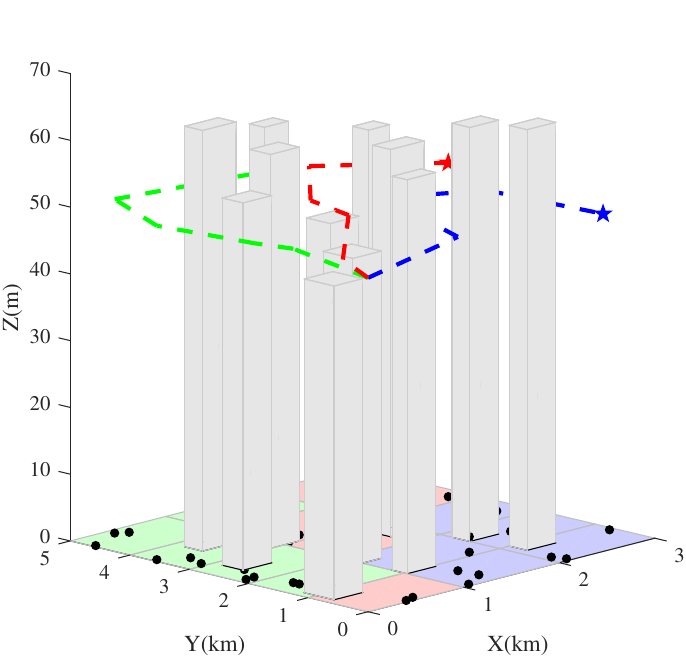}}
    \hfill
    \subfigure[Visualization of computing power.]{\includegraphics[width=0.33\textwidth]{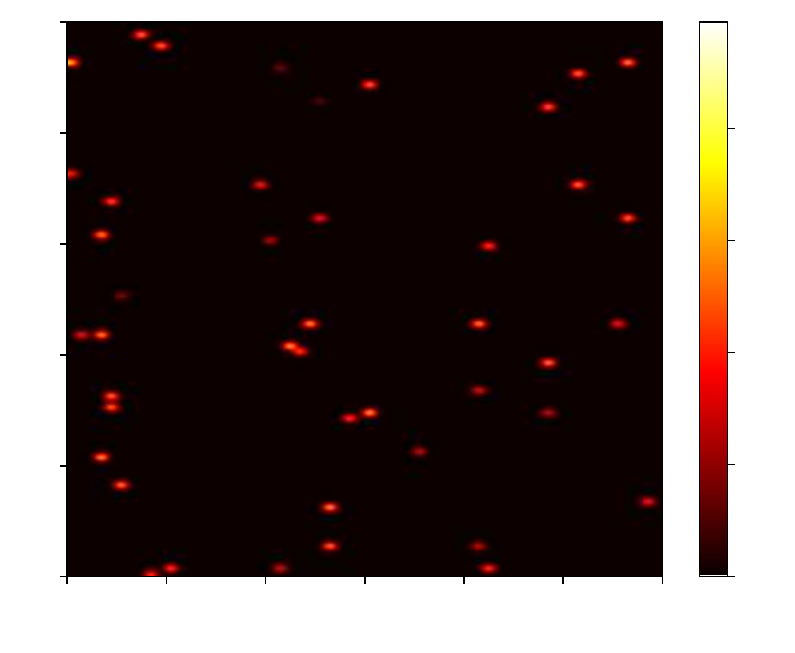}}
    \caption{Exploration optimization and GER distribution.}
    \label{fig10}
\end{figure*}

In low-altitude UAV rescue scenarios, the conditions in the rescue area are often unknown. Therefore, conducting efficient autonomous exploration of such unknown environments presents a challenge. To address this, as in Fig.~\ref{fig10}, different subareas are marked by colors. We set the number of UAVs to 3 and assigned each UAV 5 operating time slots, dividing the entire rescue area into 15 subareas for exploration. Fig.~\ref{fig10}(a) illustrates the operation of three UAVs conducting a flight within a designated area. These UAVs collaborate with one another, utilizing coordinated strategies to successfully execute the rescue mission within the area. The proposed algorithm demonstrates effective performance in exploration optimization within unknown low-altitude environments. It is capable of identifying the optimal flight trajectory in environments containing diverse obstacles, employing autonomous decision-making to navigate around these obstacles, and ultimately reaching the target destination, as in illustration Fig.~\ref{fig10}(b). This highlights the algorithm's ability to ensure autonomous exploration and trajectory planning, offering a novel approach for the widespread use of UAVs in unknown low-altitude environments.

Fig.~\ref{fig10}(c) illustrates the distribution and utilization of computing power resources of GERs in a rescue area. The main purpose of this figure is to present the remaining computation resources of each GER. From a quantitative perspective, it can be observed that UAVs can autonomously select subareas and GERs based on their coordinates and available resources, as validated in Fig.~\ref{fig10}(a). From a qualitative perspective, during rescue operations, emergency command decision-makers can quickly assess the available computation resources of GERs in the area through the visualization of their distribution and usage. This facilitates informed decision-making for the scheduling of rescue resources and the allocation of efforts.


\section{Conclusion}
\label{sec:7}
In this study, we have addressed the joint problem of task assignment and exploration optimization in low-altitude UAV rescue as a dynamic long-term optimization problem. The primary objective is to minimize the task completion time and energy consumption while ensuring system stability over an extended period. To solve this, we have first employed the Lyapunov optimization method to transform the long-term optimization problem, which includes stability constraints, into a per-slot deterministic problem. Subsequently, we have solved the UAV exploration optimization problem iteratively through the use of the Hungarian algorithm. Following this, the task assignment decisions for the UAV, including the allocation of computation resources, task offloading ratio, and GER selection, are determined by applying the HG-MADDPG method. Through extensive numerical simulations, we have demonstrated that the proposed method outperforms existing benchmark solutions in terms of performance. In future work, we plan to incorporate three-dimensional trajectory planning and network topology optimization to enhance the reliability of low-altitude UAV networks and better address the demands of low-altitude economic applications in complex environments.



\ifCLASSOPTIONcaptionsoff
  \newpage
\fi

\bibliographystyle{IEEEtran}
\bibliography{main}





\end{document}